\documentclass[11pt]{article}

\usepackage[final]{acl}
\usepackage[most]{tcolorbox}
\usepackage{xcolor}
\usepackage{float}
\definecolor{darkgray}{RGB}{60,60,60}
\definecolor{lightgray}{RGB}{245,245,245}

\usepackage{amsmath,amsfonts,bm}

\def\eqref#1{equation~\ref{#1}}

\def\1{\bm{1}}

\DeclareMathAlphabet{\mathsfit}{\encodingdefault}{\sfdefault}{m}{sl}
\SetMathAlphabet{\mathsfit}{bold}{\encodingdefault}{\sfdefault}{bx}{n}

\DeclareMathOperator*{\argmax}{arg\,max}

\usepackage{times}
\usepackage{latexsym}
\usepackage{multirow}
\usepackage{booktabs}
\usepackage{graphicx} 
\usepackage{kotex} 
\usepackage{amsmath}
\usepackage{xcolor}
\usepackage{xspace}

\usepackage[table]{xcolor} %
\definecolor{lightgray}{gray}{0.85} %

\newcolumntype{G}{>{\columncolor{gray!15}}c}

\usepackage[T1]{fontenc}

\usepackage[utf8]{inputenc}

\usepackage{microtype}

\usepackage{inconsolata}

\usepackage{graphicx}
\usepackage{hyperref}
\usepackage{url}

\newcommand{\Lasef}{\textsc{Lasef}}
\newcommand{\LASEF}{\textsc{Lasef}\xspace}

\usepackage{kotex}   %
\usepackage{comment}

\usepackage{booktabs}

\usepackage{tcolorbox}
\tcbuselibrary{breakable}
\tcbset{
  mybox/.style={
    colback=gray!5,
    colframe=black!60,
    boxrule=0.5pt,
    arc=2mm,
    outer arc=2mm,
    left=6pt,
    right=6pt,
    top=6pt,
    bottom=6pt,
    breakable
  }
}

\usepackage{graphicx}
\usepackage{wrapfig}

\usepackage{enumitem}
\usepackage[utf8]{inputenc}

\usepackage{booktabs}
\usepackage{xcolor}

\usepackage{subcaption}
\renewcommand\thesubfigure{Fig.\thefigure(\alph{subfigure})\ }

\usepackage{array}
\usepackage{makecell}
\usepackage{geometry}
\usepackage{amsmath}
\usepackage{booktabs, tabularx, makecell, threeparttable}
\usepackage{amsmath}

\usepackage[dvipsnames]{xcolor}  %

\usepackage{multirow}

\usepackage{algorithm}
\usepackage{algpseudocode}

\usepackage[table]{xcolor}
\usepackage{adjustbox}

\usepackage{booktabs}
\usepackage[table]{xcolor}
\usepackage{tabularx}
\usepackage{ragged2e}
\usepackage[table]{xcolor} %
\usepackage{colortbl}     %

\tcbuselibrary{listings} %

\usepackage{listings}

\title{Which Language Should a Skeleton Speak? Language Choices \\in Multilingual Reasoning}

\author{
    \textbf{HyeonSeok Lim$^{1}$, SeungWoo Song$^{2}$, Inho Won$^{2}$, Hoyun Song$^{2,3}$, Jihyo Kim$^{2}$} \\
    \textbf{KyungTae Lim$^{2\dagger}$} \\[4pt]
    $^{1}$ETRI \quad $^{2}$KAIST \quad $^{3}$Department of Artificial Intelligence, Dankook University \\
    \texttt{gustjrantk@etri.re.kr}, \texttt{\{sswoo,kotmul2no,jihyo.kim,ktlim\}@kaist.ac.kr}, \\
    \texttt{hoyun.song@dankook.ac.kr} \\
}
\begin{document}
\maketitle
\begingroup
\renewcommand{\thefootnote}{$\dagger$}
\footnotetext{Corresponding author.}
\endgroup

\begin{abstract}

Skeleton-based reasoning prompting is a promising training-free approach for structuring LLM reasoning, but prior work largely assumes an English-centric setting. We propose the Language-Aware Skeleton Exploration Framework (\Lasef) to study skeleton-language choice in multilingual mathematical reasoning. Across math benchmarks, model scales, and languages, we show that English skeletons yield a small positive tendency on average, most visible for smaller models and low-resource languages. However, few language-level gains remain significant after correction, and English is not universally optimal. Combining greedy decoding, multi-rollout evaluation, translation ablation, and cross-benchmark validation, we further find three patterns of skeleton-language effects: directionally consistent, evaluation- and benchmark-dependent, and asymmetric negative. These effects cannot be fully explained by generation quality alone. Overall, skeleton language is a context-dependent design variable that requires multi-level exploration. All resources are released\footnote{\url{https://github.com/lhsstn/LASEF}}.

\end{abstract}

\section{Introduction}

Recent studies on structured reasoning prompting have employed an approach~\citep{wang2023plan, khot2022decomposed}.
This approach guides a model’s reasoning process by concisely presenting the core flow of thought required for problem solving. In this paper, we refer to such explicit representations of reasoning flow as a \textit{skeleton}. A skeleton provides a high-level indication of the direction or strategy for solving a given input query, upon which more detailed reasoning and the final answer are subsequently generated~\citep{ning2023skeleton,li2024eliciting,qi2025sot}. A key property of skeleton-based reasoning is its \emph{training-free} nature, which enables its use in low-resource language scenarios.

However, when applying skeletons in multilingual settings, there exists a fundamental design challenge. A skeleton-based reasoning pipeline implicitly involves three language choices: (1) the language of the input query, (2) the language in which the skeleton is expressed, and (3) the language in which reasoning and the final answer are generated. Despite the fact that different combinations of these elements may lead the model to rely on different linguistic priors and reasoning trajectories~\citep{wang-etal-2025-language-mixing, wang2024seaeval}, prior work has not systematically explored this design space of language combinations.

Prior work on skeleton-based reasoning has largely assumed an English-centric setting~\citep{zhou2022least,qi2025sot}, focusing on how to design the structure or prompt format of skeletons to improve performance. While some studies have applied skeletons to multilingual or cross-domain scenarios~\citep{clp-emnlp2023,xlt-emnlp2023}, most of them either fix the skeleton language to English or treat language choice as a secondary factor. As a result, the crucial question of which language should be utilized for the reasoning skeleton remains insufficiently analyzed for real-world multilingual deployment.

Meanwhile, recent prior studies have reported that even the same model can exhibit different performance depending on the language in which reasoning is carried out~\citep{qi2025models}, and have shown that performing reasoning after translating the query into English can be effective~\citep{shi2022language,trans-allneed-2025naacl, zhu2024question}. These findings suggest that skeletons are not merely formal tools, but may function as a medium that activates language-specific reasoning priors formed during pretraining~\citep{etxaniz2024multilingual}. At the same time, they also imply that the effectiveness of skeletons may not be uniform across languages~\citep{qian2024large, chen2023breaking}.

In this context, we pose the following research question: \textit{Under which linguistic conditions does skeleton-based reasoning operate effectively, and how do different combinations of input language, skeleton language, and reasoning/answer language affect reasoning performance? In particular, how do these effects differ in low-resource languages?} This question goes beyond asking ``which skeleton is effective,'' and instead addresses a more fundamental design issue: which language should be used for the skeleton.

To investigate this question, we propose \textit{Language-Aware Skeleton Exploration}. Here, exploration does not involve any new algorithmic learning or parameter updates; instead, it refers to an empirical approach that systematically analyzes performance variations across different language combinations while keeping the skeleton structure and format fixed, with the goal of identifying effective language configurations. This setup allows us to disentangle the impact of skeleton design itself from that of language choice.

Our study focuses on mathematical reasoning, a domain where multilingual performance gaps are known to be significant due to limited reasoning capabilities in non-English languages~\citep{qi2025models}. We conduct experiments across benchmarks of varying difficulty levels, models of different scales, and a wide range of language settings spanning high-resource to low-resource languages. Through this comprehensive evaluation, we systematically analyze how the effectiveness of skeletons varies across languages and conditions. Building on these findings, we derive practical insights for the reliable use of skeleton-based reasoning in multilingual environments. Our results answer the above question conditionally rather than universally: instead of prescribing a single optimal skeleton language, we characterize when skeleton-language choices help or hurt and how consistently these effects generalize across decoding protocols and benchmarks. In summary, the contributions of this paper are as follows:

\begin{itemize}
    \item We formalize the combination of input, skeleton, and reasoning languages as a core design variable in multilingual skeleton-based reasoning, and propose a training-free framework (\Lasef) to analyze this design space.
    
    \item Combining greedy decoding and multi-rollout evaluation across math benchmarks of varying difficulty, we demonstrate the non-uniform effectiveness of skeleton languages and assess the directional consistency of the observed effects across decoding protocols and benchmarks.
    
    \item We identify three distinct patterns of skeleton-language effects: directionally consistent, evaluation- and benchmark-dependent, and asymmetric negative. These patterns are supported by a translation ablation that disentangles skeleton generation quality from cross-lingual alignment.
\end{itemize}

\section{Related Work}
\paragraph{English-Pivot Reasoning}
While Chain-of-Thought (CoT) prompting~\citep{cot-neurips2022} has enhanced LLM reasoning, a significant performance gap persists in non-English contexts, particularly in mathematical tasks~\citep{shi2022language, mega-emnlp2023, wang2025polymath}. 
To mitigate this disparity, English-pivot strategies that translate inputs or reasoning processes into English have been proposed to leverage the models' strong English priors~\citep{clp-emnlp2023, xlt-emnlp2023, trans-allneed-2025naacl, etxaniz2024multilingual}. However, this approach often sacrifices interpretability for non-English users by generating final answers in English, thereby limiting practical utility in native-language applications~\citep{multilingual-prompting-emnlp2025,lee2025controlling,qi2025models}.

\paragraph{Structured Reasoning} 
Research on structure-aware prompting has evolved from Zero-shot CoT~\citep{kojima2022large} to more sophisticated approaches such as Skeleton-of-Thought~\citep{ning2023skeleton}, Plan-and-Solve~\citep{wang2023plan}, and Program-of-Thought~\citep{chen2022program}. These methods have demonstrated that explicitly structuring reasoning enhances logical consistency~\citep{zhou2022least,yao2023tree,khot2022decomposed}. While efforts have been made to extend structured reasoning to multilingual contexts through cross-lingual instruction tuning~\citep{chai2025xcot,lai2024mcot}, prior work has predominantly fixed the skeleton language to English, overlooking the impact of language combinations between the structural guide and the final response~\citep{qi2025sot,li2024eliciting}.

In this study, we refer to such structured guides as a \textit{skeleton} and adopt the inference structure of \citet{qi2025sot} as a baseline to ensure experimental control. By fixing the structure and manipulating only the skeleton's language, we systematically analyze how language configuration affects multilingual reasoning performance.

\begin{figure*}[t]
    \centering
    \includegraphics[width=\textwidth]{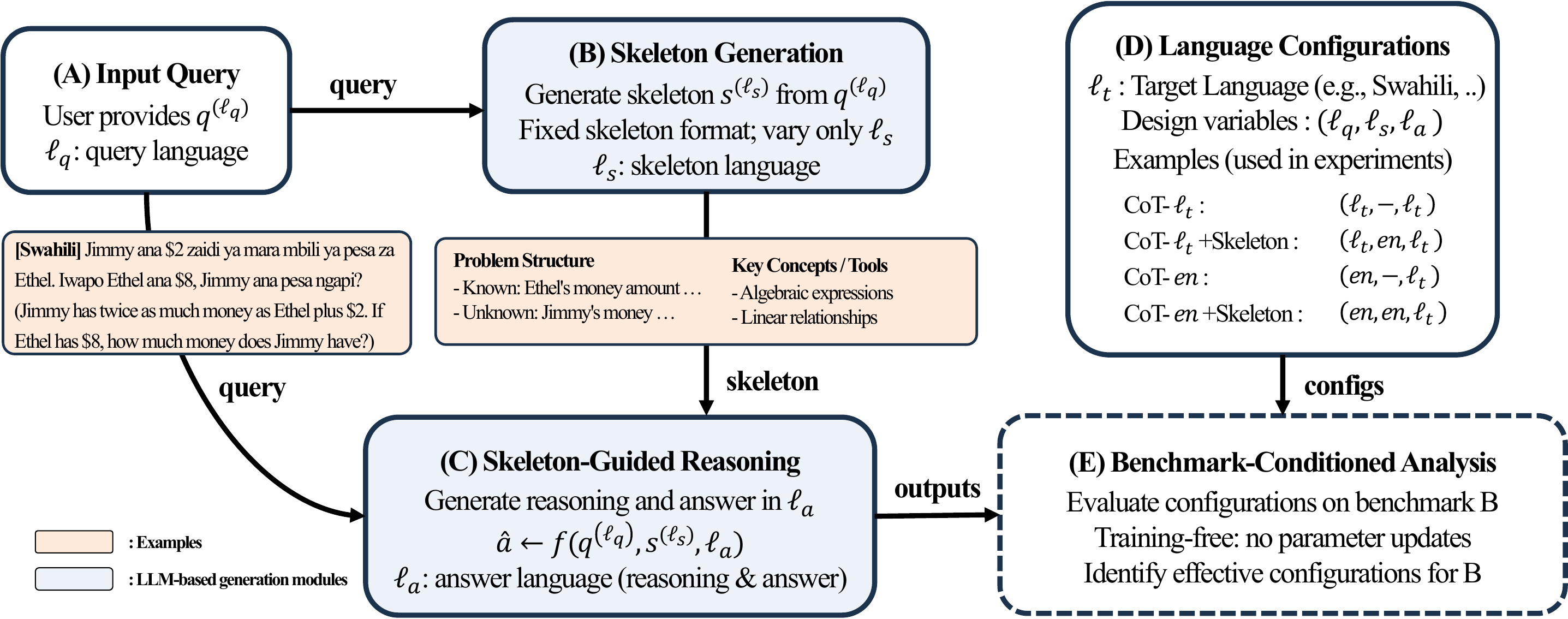}
    \caption{\small Overview of the Skeleton-Guided Reasoning Framework. (A) The process begins with an input query $q^{(\ell_q)}$. (B) A structured skeleton $s^{(\ell_s)}$ is generated to abstract the problem logic. (C) The final reasoning and answer $\hat{a}$ are produced in the answer language $\ell_a$, conditioned on the query and skeleton. (D) We define the language configuration space as $(\ell_q, \ell_s, \ell_a)$ and compare four primary strategies: standard CoT and skeleton-guided approaches, applied in both the target language ($\ell_t$) and English ($en$). (E) Effective configurations are identified through training-free benchmark analysis.
}
\label{fig:figure1}
    \vspace{-0.15in}
\end{figure*}

\section{Language-Aware Skeleton Exploration Framework}

This study investigates how the effectiveness of skeleton-based reasoning varies depending on language choice and evaluation contexts within a multilingual environment. To this end, we propose the \emph{Language-Aware Skeleton Exploration Framework (\Lasef)}, as illustrated in Fig.~\ref{fig:figure1}, which systematically analyzes skeleton utilization by varying language configurations while keeping the structure and format of the skeleton fixed. Here, ``exploration'' does not refer to new training or algorithmic search, but rather to an empirical analysis process designed to identify effective language configurations within a given evaluation environment.

\subsection{Problem Setup and Language Dimensions}

Standard language model inference typically involves a query $q$ and a corresponding answer $a$, where the query is presented in a specific language $\ell_q$. In skeleton-based reasoning, the model first generates a skeleton $s$, summarizing the core structure required for problem-solving based on the query, and subsequently uses it to generate the detailed reasoning process and the final answer. This process entails three distinct language choices given a user's target language $\ell_t$: (1) the language of the query, $\ell_q$; (2) the language of the skeleton, $\ell_s$; and (3) the language in which the detailed reasoning and final answer are generated, $\ell_a$.

These three languages may be identical or distinct; depending on their combination, the linguistic priors and reasoning paths elicited by the model can vary. Thus, a single skeleton-based reasoning process is characterized by the language combination $(\ell_q, \ell_s, \ell_a)$. \LASEF focuses on isolating and analyzing the impact of these language configurations on reasoning performance while fixing the skeleton generation method and structure. Under this formulation, the skeleton-based reasoning process can be expressed as:
\[
\hat{a} = f\big(q^{(\ell_q)}, s^{(\ell_s)}, \ell_a\big),
\]

where $f(\cdot)$ represents the inference function of the pre-trained language model, and $s^{(\ell_s)}$ denotes the skeleton expressed in language $\ell_s$.

Notably, the effective language combination for skeleton utilization is not universally fixed but may vary depending on the benchmark or domain used for evaluation. Given a specific benchmark $B$ consisting of query-answer pairs $(\{q,a\} \in B)$, \LASEF provides an analytical framework to identify relatively effective skeleton language configurations within that evaluation environment by comparing performance variations under different language combinations $(\ell_q, \ell_s, \ell_a)$.

This problem setup assumes a \emph{training-free} environment involving no additional training or parameter updates. The following section describes the specific language configurations and skeleton usage methods considered in this study.

\subsection{Skeleton Construction and Usage}

The skeletons used in this study are intermediate representations that concisely reveal the \emph{reasoning structure} for problem-solving, rather than presenting core answers or calculation procedures. In other words, the skeleton provides a minimal structural guide for the reasoning process to follow, avoiding specific instructions for the actual solution steps. This perspective conceptually aligns with existing research interpreting skeletons as structural representations that organize reasoning, rather than as summaries or hints for the solution.

Accordingly, our skeletons maintain a very short and abstract form, excluding specific calculation steps or answers%
. To ensure this, we consistently apply the following principle during skeleton generation: \emph{``Provide only the essential structural characteristics of the problem. Describe what the problem structure is, never how to solve it.''}

Fig.~\ref{fig:figure1} presents a representative example of a skeleton generated in English ($\ell_s=en$) for a Swahili query ($\ell_q=sw$). Adhering to our proposed criteria, skeletons maintain a concise length of approximately 120 tokens on average regardless of the language, devoid of direct answers or core solution steps. We provide a detailed quantitative analysis of skeleton length and content distribution, along with the complete generation rules and additional examples, in Appendix~\ref{Appendix:about-skeleton}.
\section{Experiments}
In this section, we systematically validate the effectiveness of the proposed \emph{Language-Aware Skeleton Exploration Framework (\Lasef)} across diverse settings. The experiments are designed to analyze performance by varying model scale, problem difficulty, and language resource levels, while keeping the structure and generation method of the skeleton fixed. Through this, we analyze (1) whether the effectiveness of skeleton-based reasoning is limited to specific conditions, and (2) how it varies according to language and model characteristics.

\subsection{Experimental Setup}
\paragraph{Models based on Scale.} 
To analyze the impact of model scale, we select the \textsc{Qwen2.5} series~\citep{qwen2025qwen25technicalreport} as our base models, following prior research~\citep{qi2025sot}. These models are chosen for their superior multilingual capabilities as general-purpose models, rather than being specialized solely for reasoning. We evaluate 7B, 14B, and 72B models to examine how skeleton guidance scales. For brevity, we refer to the \textsc{Qwen2.5-Instruct} models by their sizes (e.g., \textsc{Qwen2.5-14B}). Results for the \textsc{Llama~3.1} series~\citep{grattafiori2024llama3herdmodels} are provided in Appendix~\ref{Appendix:Llama-results}.

\paragraph{Benchmarks by Difficulty.} 
We categorize our benchmarks by difficulty to assess skeleton effectiveness across task complexities. We employ MGSM~\citep{shi2022language} for elementary arithmetic, MATH-500~\citep{lightman2023let} for intermediate-level problems, and PolyMath~\citep{wang2025polymath} for advanced reasoning tasks requiring complex structural analysis.

\paragraph{Target Languages based on Resource Levels.} To evaluate the impact of skeletons according to language resource levels, we included diverse target languages, from high-resource to low-resource. High-resource languages were selected based on their sufficient exposure during large-scale pre-training, while low-resource languages were chosen based on the relative scarcity of training data, which often leads to inconsistent reasoning performance in multilingual LLMs. This allows us to analyze whether skeletons play different roles depending on the resource level of the language.

\paragraph{Evaluation Protocol.} 
We use Exact Match (EM) as the evaluation metric, restricted to samples satisfying $\ell_a=\ell_t$ to ensure multilingual reasoning validity; reported scores are therefore \emph{conditional reasoning accuracies} on language-compliant samples. Alongside greedy decoding, we conduct multi-rollout evaluation ($T{=}0.7$, $N{=}5$) to assess the stability of skeleton-language effects. Details on language identification, filtering and retention rates, and pairing are in Appendix~\ref{Appendix:Evaluation-setup}, and unfiltered results are in Appendix~\ref{app:lowresource-full}.

\begin{table*}[!t]
\centering
\footnotesize
\resizebox{\textwidth}{!}{
\begin{tabular}{lcccccc>{\columncolor{yellow!15}}c|ccccc>{\columncolor{yellow!15}}c|ccccc>{\columncolor{yellow!15}}c}
\toprule
\multirow{2}{*}{\centering\textbf{Method}} & \textbf{Language} & \multicolumn{6}{c}{\textbf{MGSM}} & \multicolumn{6}{c}{\textbf{MATH-500}} & \multicolumn{6}{c}{\textbf{PolyMath}} \\
\cmidrule(lr){2-2} \cmidrule(lr){3-8} \cmidrule(lr){9-14} \cmidrule(lr){15-20}
& \textbf{($\ell_q,\ell_s,\ell_a$)} & \textbf{zh} & \textbf{es} & \textbf{ko} & \textbf{th} & \textbf{sw} & \textbf{AVG.} & \textbf{zh} & \textbf{es} & \textbf{ko} & \textbf{th} & \textbf{sw} & \textbf{AVG.} & \textbf{zh} & \textbf{es} & \textbf{ko} & \textbf{th} & \textbf{sw} & \textbf{AVG.} \\
\midrule
\rowcolor{gray!10} \multicolumn{20}{c}{\textbf{\textit{Qwen2.5-7B-Instruct}}} \\
\midrule
\textsc{CoT-$\ell_t$} & $\ell_t, - , \ell_t$ &
79.9 & 80.7 & 69.7 & 80.2 & 16.0 & 65.3 & 
56.4 & 57.5 & 47.8 & 49.8 & 14.2 & 45.1 & 
30.8 & 31.0 & 26.4 & 26.1 & 5.1  & 23.9 \\
+ \textsc{Skeleton} & $\ell_t, en , \ell_t$ &
80.7 & 83.8 & 71.0 & 79.0 & 22.9 & 67.5 & 
55.4 & 57.7 & 49.3 & 51.9 & 20.4 & 46.9 & 
30.6 & 32.1 & 28.0 & 27.6 & 7.4  & 25.1 \\
\textsc{CoT-$en^{\dagger}$} & $en, - , \ell_t$ &
82.3 & 84.3 & 74.5 & 78.1 & 63.1 & 76.5 & 
56.0 & 57.4 & 51.6 & 49.5 & 24.4 & 47.8 & 
29.7 & 32.1 & 27.5 & 29.8 & 20.3 & 27.9 \\
+ \textsc{Skeleton} & $en, en , \ell_t$ &
82.3 & 85.9 & 73.2 & 78.9 & 71.7 & 78.4 & 
57.4 & 58.2 & 52.3 & 53.8 & 29.8 & 50.3 & 
32.0 & 34.4 & 27.7 & 30.4 & 19.9 & 28.9 \\

\midrule
\rowcolor{gray!10} \multicolumn{20}{c}{\textbf{\textit{Qwen2.5-14B-Instruct}}} \\
\midrule
\textsc{CoT-$\ell_t$} & $\ell_t, - , \ell_t$ &
88.8 & 90.0 & 77.9 & 87.6 & 45.0 & 77.9 &  
58.3 & 62.6 & 54.7 & 57.4 & 30.1 & 52.6 & 
31.8 & 36.7 & 31.6 & 30.9 & 14.7 & 29.1 \\
+ \textsc{Skeleton} & $\ell_t, en , \ell_t$ &
88.0 & 90.0 & 80.7 & 85.9 & 50.7 & 79.1 & 
59.4 & 64.0 & 56.9 & 59.9 & 35.2 & 55.1 & 
34.1 & 36.9 & 32.2 & 33.3 & 15.0 & 30.3 \\
\textsc{CoT-$en^{\dagger}$} & $en, - , \ell_t$ & 
83.1 & 88.3 & 81.2 & 84.8 & 74.5 & 82.4 & 
58.8 & 60.2 & 53.8 & 55.1 & 39.1 & 53.4 & 
32.3 & 36.5 & 30.1 & 32.0 & 24.0 & 31.0 \\
+ \textsc{Skeleton} & $en, en , \ell_t$ & 
86.8 & 87.9 & 82.5 & 84.8 & 74.5 & 83.3 & 
59.6 & 62.2 & 55.2 & 55.8 & 40.4 & 54.6 &
33.7 & 35.4 & 32.2 & 32.0 & 26.9 & 32.0 \\

\midrule
\rowcolor{gray!10} \multicolumn{20}{c}{\textbf{\textit{Qwen2.5-72B-Instruct}}} \\
\midrule
\textsc{CoT-$\ell_t$} & $\ell_t, - , \ell_t$ &
88.3 & 91.6 & 83.9 & 90.8 & 68.0 & 84.5 & 
63.1 & 63.8 & 63.2 & 61.6 & 46.3 & 59.6 & 
36.1 & 37.5 & 34.5 & 34.7 & 31.0 & 34.8 \\
+ \textsc{Skeleton} & $\ell_t, en , \ell_t$ &
88.0 & 89.2 & 84.7 & 89.6 & 70.1 & 84.3 & 
65.8 & 63.8 & 62.4 & 63.4 & 54.3 & 61.9 & 
37.1 & 37.5 & 37.0 & 35.0 & 32.0 & 35.7 \\
\textsc{CoT-$en^{\dagger}$} & $en, - , \ell_t$ & 
85.5 & 91.5 & 82.7 & 86.3 & 85.5 & 86.3 & 
62.2 & 64.4 & 63.1 & 61.7 & 56.3 & 61.5 & 
34.2 & 35.8 & 35.3 & 35.8 & 40.6 & 36.3 \\
+ \textsc{Skeleton} & $en, en , \ell_t$ & 
86.8 & 90.7 & 83.1 & 86.7 & 86.7 & 86.8 & 
64.2 & 64.0 & 61.1 & 62.5 & 60.0 & 62.4 & 
34.0 & 37.9 & 36.2 & 33.1 & 39.3 & 36.1 \\
\bottomrule
\end{tabular}%
}
\caption{\small Results on MGSM, MATH-500, and PolyMath. $(\ell_q,\ell_s,\ell_a)$ denote the query, skeleton, and answer languages, respectively. CoT-$en^{\dagger}$ indicates queries translated into English using Google Translate, and the skeleton language is fixed to English unless otherwise noted. `AVG.' denotes the average over five languages (zh, es, ko, th, sw). Greedy decoding throughout.}
\label{tab:combined-main-table}
    \vspace{-0.15in}
\end{table*}
\subsection{Baseline Configurations}
To analyze the effectiveness of \Lasef, we define the language combination $(\ell_q, \ell_s, \ell_a)$ for a given target language $\ell_t$, and establish an appropriate comparative baseline. Since skeletons function as structural guides for improving reasoning performance, we adopt a Chain-of-Thought (CoT)-based setting as the primary baseline~\citep{cot-neurips2022}, following prior work on multilingual reasoning~\citep{xlt-emnlp2023,clp-emnlp2023}.

\paragraph{Target-Language CoT (\textsc{CoT-$\ell_t$}).} In this setting, the input query is given in the target language, and both reasoning and the answer are generated in the same language $(\ell_q=\ell_a=\ell_t)$. We use a standard CoT prompt (``Let’s think step by step, Respond in {$\ell_t$}'') without applying a skeleton. This represents the most natural usage scenario where users verify both the query and the reasoning process in their native language.

\paragraph{English-Pivot CoT (\textsc{CoT-$en$}).} In this setting, the input query is translated into English $(\ell_q=\text{en})$, after which the reasoning and answer are generated in the target language $\ell_t$. This baseline is essential for two reasons. First, comparing performance based on English-translated queries is a widely used practice in existing multilingual reasoning studies~\citep{clp-emnlp2023,xlt-emnlp2023,trans-allneed-2025naacl}, representing the upper-bound performance of LLMs pre-trained primarily on English. Second, even in real-world non-English user environments, since users already understand the meaning of the query, internally converting the query to English does not hinder meaning transmission. Thus, \textsc{CoT-$en$} serves as a rational baseline to evaluate whether performance gains via translation are achievable without structural guidance.

\paragraph{Skeleton Baseline Language Configuration.} 
English skeletons are used in Tab.~\ref{tab:combined-main-table} and \ref{tab:low-resource-mgsm}, while Fig.~\ref{fig:cross-lingual-skeleton} and \ref{fig:translation_scatter} expands to Chinese, Spanish, Russian, Korean, and Thai.

\subsection{Performance across Diverse Dimensions}
Tab.~\ref{tab:combined-main-table} compares \LASEF in an English skeleton setting $(\ell_q, en, \ell_a)$ against baselines across three benchmarks: MGSM, MATH-500, and PolyMath.

\paragraph{Bridging the Resource Gap}
Language resource levels significantly influence the impact of English skeletons. Gains are substantially larger for low-resource languages like Swahili~(sw) compared to high-resource languages like Chinese~(zh) and Spanish~(es). For instance, with \textsc{Qwen2.5-7B} on MGSM (CoT-$en^{\dagger}$), Swahili performance rose from 63.1 to 71.7 (+8.6), while Chinese and Spanish showed marginal gains (+0.0 and +1.6). This trend persists across scales; with \textsc{Qwen2.5-72B} on MATH-500, Swahili again saw the highest improvement (+3.7). These results indicate that when low-resource languages lack sufficient reasoning stability, English skeletons serve as an essential scaffold to stabilize the process.

\paragraph{Does Scaling Always Guarantee Improvement?}
With respect to model scale, we observe that smaller models benefit more from English skeletons, while the effect diminishes or even becomes negative as model size increases. \textsc{Qwen2.5-7B} and \textsc{14B} exhibit performance improvements after applying skeletons across all benchmarks (+0.9 to 2.5). In contrast, \textsc{Qwen2.5-72B} shows an average performance drop of 0.2 after skeleton application under the \textsc{CoT}-$\ell_t$ setting on MGSM and the \textsc{CoT}-$en^{\dagger}$ setting on PolyMath. This suggests that for large-scale models, externally imposed English skeletons may act as an unnecessary constraint rather than a helpful inductive bias.

\paragraph{Does Benchmark Difficulty Limit the Gains?}
Examining performance variations across benchmark difficulty levels, we observe that the contribution of skeletons is larger for easier benchmarks. Using \textsc{Qwen2.5-7B} as the reference model, skeletons yield a performance gain of +2.2 on MGSM, the easiest benchmark, compared to +1.8 on MATH-500, which has medium difficulty, and +1.2 on PolyMath, the most challenging benchmark. This trend suggests that as problem difficulty increases, simple skeleton generation alone becomes insufficient, as harder problems require more sophisticated and precise computational reasoning beyond what skeletons can provide.

\paragraph{Summary of Results.}

Results show that English skeletons' benefits depend on model scale, task difficulty, and language resources. Gains are most pronounced for smaller models, easier tasks, and low-resource languages. Notably, the CoT-\textsc{$en^{\dagger}$} setting consistently outperforms CoT\textsc{-$\ell_t$}, reaffirming the strong English prior of LLMs. This underscores translation as a critical baseline. In low-resource settings, most of the improvement comes from translation rather than from skeletons (+24.07 vs.\ +2.02 pp in Tab.~\ref{tab:low-resource-mgsm}).

\begin{table}[!t]
\centering
\resizebox{\columnwidth}{!}{
\begin{tabular}{lrrrr}
\toprule
\multirow{2}{*}{\textbf{Language}} &
\multicolumn{2}{c}{\textbf{$\ell_t,en,\ell_t$}} &
\multicolumn{2}{c}{\textbf{$en^{\dagger},en,\ell_t$}}\\
\cmidrule(lr){2-3}\cmidrule(lr){4-5}
 & \textsc{CoT-$l_t$} & +\textsc{Skeleton} & \textsc{CoT-$en^{\dagger}$} & +\textsc{Skeleton} \\
\midrule
\rowcolor{gray!10}\multicolumn{5}{c}{\textbf{\textit{Qwen2.5-7B-Instruct}}}\\
\midrule
kk & 40.08 & 51.42 (\textcolor{red}{+11.34}) & 62.82 & 72.22 (\textcolor{red}{+9.40}) \\
ky & 33.33 & 37.86 (\textcolor{red}{+4.53}) & 61.92 & 65.69 (\textcolor{red}{+3.77}) \\
mn & 27.16 & 33.33 (\textcolor{red}{+6.17}) & 66.96 & 64.78 (\textcolor{blue}{-2.18}) \\
ug & 27.13 & 37.65 (\textcolor{red}{+10.52}) & 57.14 & 60.37 (\textcolor{red}{+3.23}) \\
hy & 50.81 & 56.05 (\textcolor{red}{+5.24}) & 64.49 & 73.88 (\textcolor{red}{+9.39}) \\
lo & 25.51 & 23.48 (\textcolor{blue}{-2.03}) & 48.54 & 48.95 (\textcolor{red}{+0.41}) \\
eu & 20.76 & 19.92 (\textcolor{blue}{-0.84}) & 65.62 & 74.55 (\textcolor{red}{+8.93}) \\
mt & 25.65 & 20.00 (\textcolor{blue}{-5.65}) & 62.11 & 58.59 (\textcolor{blue}{-3.52}) \\
am & 8.23  & 10.29 (\textcolor{red}{+2.06})  & 39.63 & 32.93 (\textcolor{blue}{-6.70}) \\
ne & 54.29 & 52.65 (\textcolor{blue}{-1.64}) & 74.58 & 78.81 (\textcolor{red}{+4.23}) \\
si & 10.16 & 12.20 (\textcolor{red}{+2.04})  & 37.14 & 35.92 (\textcolor{blue}{-1.22}) \\
ps & 26.46 & 26.91 (\textcolor{red}{+0.45}) & 37.62 & 47.03 (\textcolor{red}{+9.41}) \\
sd & 20.27 & 22.07 (\textcolor{red}{+1.80}) & 64.81 & 68.52 (\textcolor{red}{+3.71}) \\
pa & 51.41 & 52.21 (\textcolor{red}{+0.80}) & 55.87 & 52.63 (\textcolor{blue}{-3.24}) \\
gu & 44.35 & 47.98 (\textcolor{red}{+3.63}) & 59.84 & 60.64 (\textcolor{red}{+0.80}) \\
mr & 45.34 & 46.96 (\textcolor{red}{+1.62}) & 75.41 & 74.18 (\textcolor{blue}{-1.23}) \\
or & 44.35 & 43.95 (\textcolor{blue}{-0.40}) & 36.84 & 38.06 (\textcolor{red}{+1.22}) \\
ta & 47.79 & 49.80 (\textcolor{red}{+2.01}) & 69.48 & 69.08 (\textcolor{blue}{-0.40}) \\
ml & 42.74 & 46.37 (\textcolor{red}{+3.63}) & 62.25 & 64.66 (\textcolor{red}{+2.41}) \\
\midrule
\textbf{AVG.}
 & 33.99 & 36.37 (\textcolor{red}{+2.38})
 & 58.06 & 60.08 (\textcolor{red}{+2.02}) \\
\bottomrule
\end{tabular}
}
\caption{\small Performance comparison on 19 low-resource languages (MGSM) using \textsc{Qwen2.5-7B} with English Skeleton. CoT-$en^{\dagger}$: English-translated queries via Google Translate. Parentheses show $\Delta$ gains.}
\label{tab:low-resource-mgsm}
    \vspace{-0.15in}
\end{table}

\subsection{Effect of English Skeletons in Low-Resource Languages} \label{sec:robust_lowresource}
Previous analysis confirms that English skeletons are most effective for smaller models, easier benchmarks, and low-resource languages. In this section, we validate this finding by evaluating \textsc{Qwen2.5-7B} on 19 low-resource MGSM languages. This section examines the consistency of skeleton-based CoT improvements under these conditions. The criteria for language selection and our translation methodology are detailed in Appendix~\ref{appendix:mgsm-translation}.

Tab.~\ref{tab:low-resource-mgsm} summarizes the performance changes under the target-language input~($\ell_t$) and English-translated input~($en^{\dagger}$) settings. English skeletons show a small positive tendency on average: applying them improves 13 of the 19 languages, with mean gains of +2.38 in the CoT-$\ell_t$ setting and +2.02 in the CoT-$en^{\dagger}$ setting. However, after FDR correction, statistically significant language-level improvements are observed only for Kazakh~(kk) and Uyghur~(ug); pooling the 19 languages at the item level, the overall effect remains significant (Appendix~\ref{app:stats}).

First, under the Native Input~(CoT-$\ell_t$) setting, skeletons substantially unlock the model’s latent reasoning capacity for certain languages, such as Kazakh~(kk) (+11.34) and Uyghur~(ug) (+10.52). However, there also exist cases where performance degrades, such as Maltese~(mt) (−5.65), indicating a high variance in language-specific effects.

Second, under the (CoT-$en^{\dagger}$) setting, the effect of English skeletons becomes considerably more consistent in direction. Notably, query translation alone leads to a dramatic increase in average performance, from 33.99 to 58.06. When English skeletons are incorporated, an additional performance gain of +2.02 is observed. In particular, languages such as Armenian~(hy)~(+9.39), Basque~(eu)~(+8.93), and Pashto (ps)~(+9.41) exhibit substantial improvements when structure in the form of skeletons is added on top of translated queries.

Overall, in low-resource language settings, query translation provides the dominant improvement, and English skeletons contribute a smaller additional gain by structuring the reasoning process. This suggests that skeleton-based CoT can serve as a useful supplementary reasoning aid---rather than the main driver of improvement---even in extremely low-resource language scenarios.

\subsection{Is English the Only Solution for Low-Resource Languages?}

\begin{figure*}[!ht]
    \centering
    \captionsetup[subfigure]{labelformat=simple, labelsep=space}
    \renewcommand\thesubfigure{(\alph{subfigure})}
    \begin{subfigure}[t]{0.48\linewidth}
        \centering
        \includegraphics[width=\linewidth]{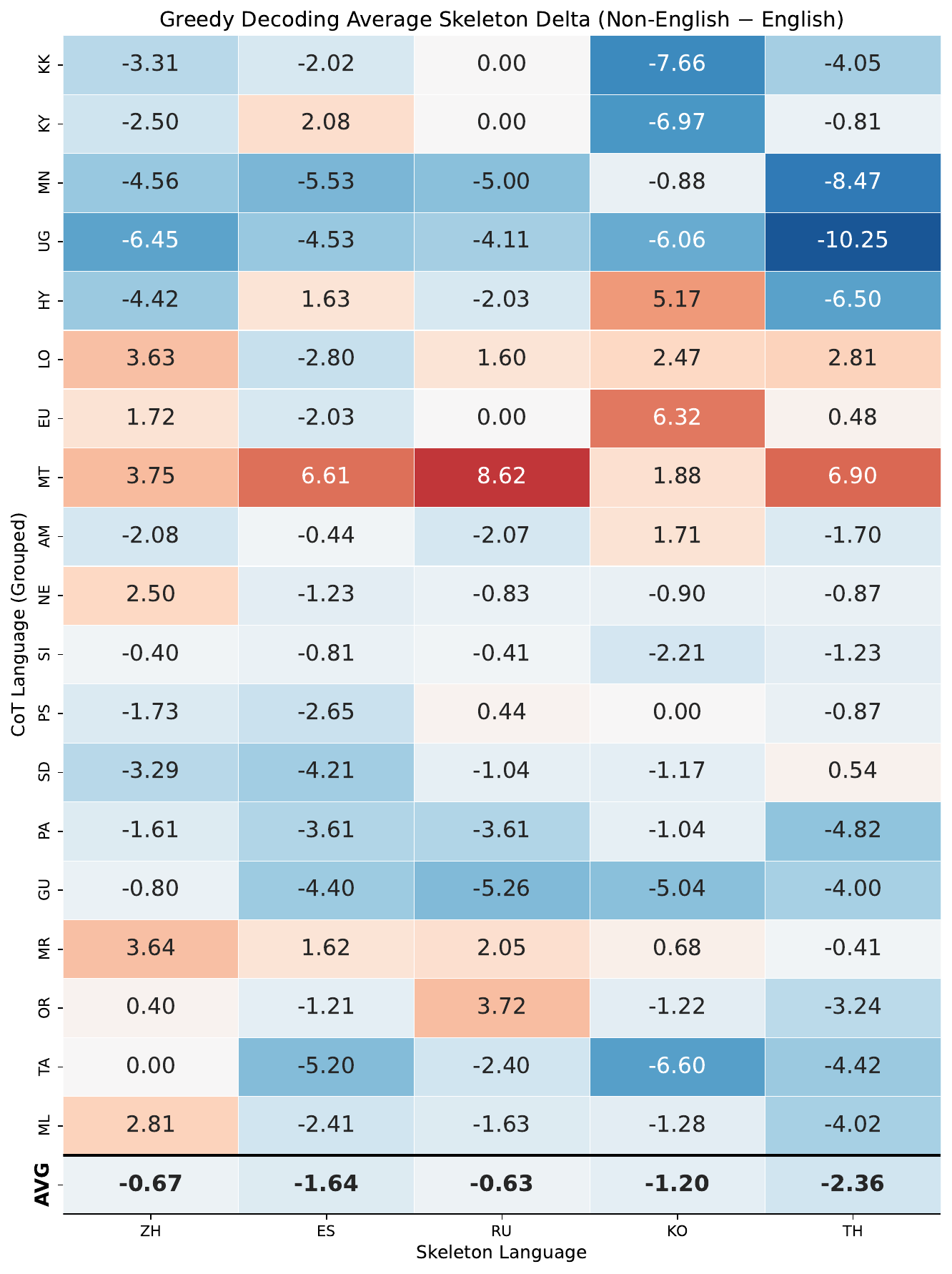}
        \caption{Greedy Decoding (single rollout)}
        \label{fig:heatmap-greedy}
    \end{subfigure}
    \hfill
    \begin{subfigure}[t]{0.48\linewidth}
        \centering
        \includegraphics[width=\linewidth]{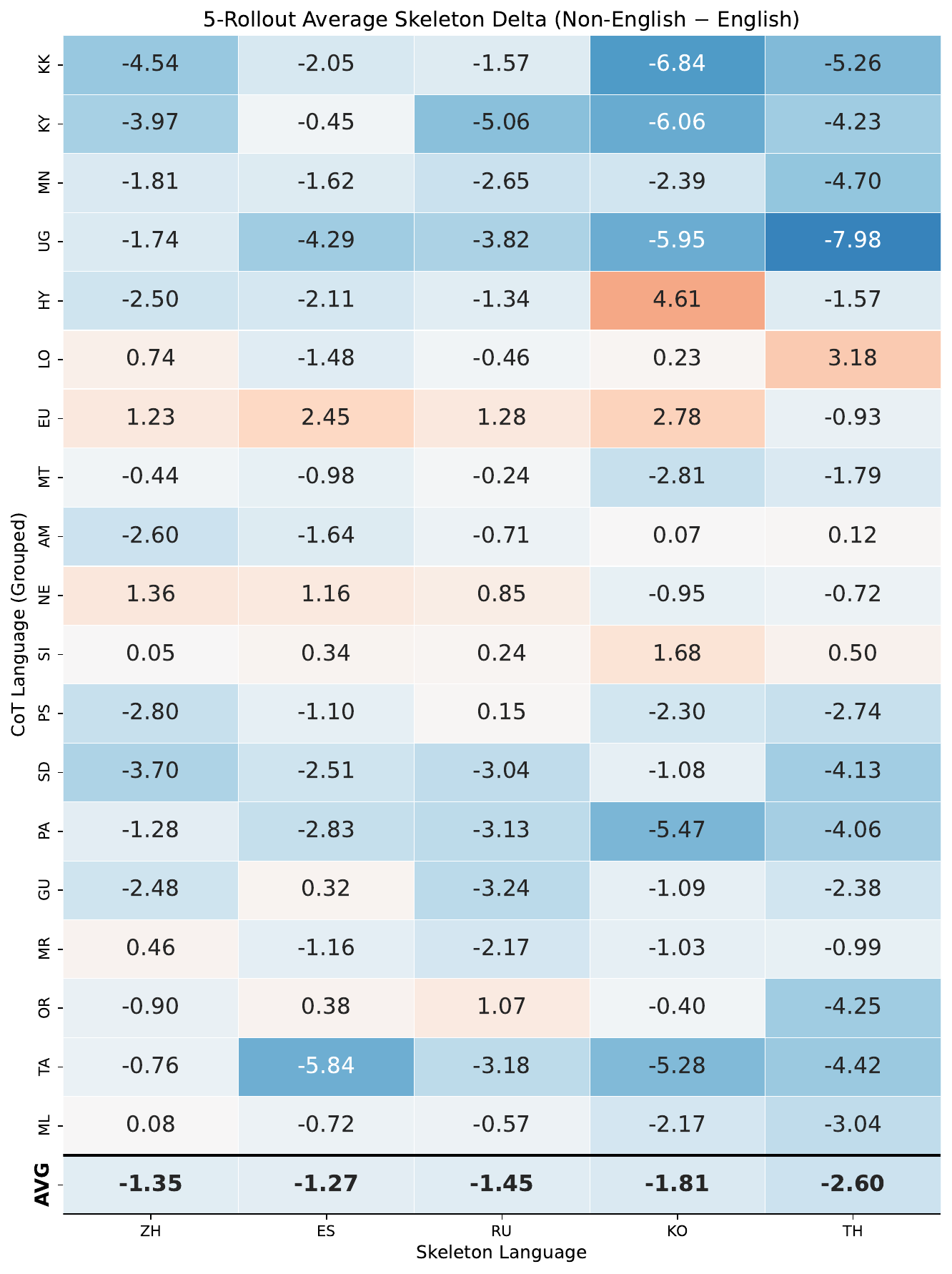}
        \caption{Multi-rollout ($T{=}0.7$, $N{=}5$)}
        \label{fig:heatmap-rollout5}
    \end{subfigure}
    \caption{\small Performance differences (Non-English $-$ English) on MGSM under CoT-$\ell_t$ with \textsc{Qwen2.5-7B}. The English skeleton is replaced with five Non-English skeletons (ZH, ES, RU, KO, TH); the same skeleton set is used in both panels. \textbf{(a)} Greedy decoding (single rollout). \textbf{(b)} Multi-rollout sampling ($T{=}0.7$, $N{=}5$). The corresponding English-skeleton reference values (English skeleton vs.\ no skeleton per language) are reported in Tab.~\ref{tab:low-resource-mgsm}.}
    \label{fig:cross-lingual-skeleton}
    \vspace{-0.15in}
\end{figure*}
\label{sec:low-resource-nonen}
In Sec.~\ref{sec:robust_lowresource}, we showed that English skeletons generally serve as effective reasoning aids. However, the \textsc{CoT-$\ell_t$} results in Tab.~\ref{tab:low-resource-mgsm} indicate that English skeletons are not always beneficial for all low-resource languages. In particular, five out of the 19 languages (lo, eu, mt, ne, and or) show performance degradation when English skeletons are applied (the 19 languages are those of the 34 MGSM low-resource languages that pass the sample filter; see Appendix~\ref{Appendix:Evaluation-protocol}).

In this section, we extend \LASEF to five non-English skeletons (zh, es, ru, ko, and th) and analyze how the effect of skeleton language choice varies across target languages. We distinguish surface \textit{affinity} (language family, script, and geographic proximity) from structural \textit{alignment} (typology and word order), and examine the stability of the observed effects using both greedy decoding and multi-rollout sampling.

\paragraph{General Trends: English as a Stable Anchor.}
Fig.~\ref{fig:cross-lingual-skeleton} shows the performance differences between non-English and English skeletons under the \textsc{CoT-$\ell_t$} setting with \textsc{Qwen2.5-7B}, using greedy decoding and multi-rollout evaluation. Overall, applying non-English skeletons uniformly leads to an average performance decrease across languages, ranging from 0.63 to 2.60 points. This result reaffirms that English serves as a stable reasoning skeleton for most languages. However, when examining individual languages, notable exceptions emerge, particularly for languages on which the English skeleton had a negative effect.

\paragraph{Directionally Consistent Cross-Lingual Effects.}
As shown in Tab.~\ref{tab:low-resource-mgsm}, some languages that experience performance degradation with English skeletons exhibit clear improvements when paired with alternative skeletons that share affinity or alignment with the target language. Lao~(lo) suffers a performance drop with the English skeleton ($-2.03$), but shows a $+2.81$ improvement with a Thai skeleton, which shares surface affinity through the Tai--Kadai language family (Fig.~\ref{fig:heatmap-greedy}). Similarly, Basque~(eu) and Armenian~(hy) achieve gains of $+6.32$ and $+5.17$, respectively, when paired with a Korean skeleton, which shares their structural alignment (agglutinative morphology and SOV word order). Crucially, \textit{the sign of these effects is preserved under multi-rollout evaluation} (Fig.~\ref{fig:heatmap-rollout5}): hy--ko slightly decreases ($+5.17 \to +4.61$), lo--th is preserved ($+2.81 \to +3.18$), and eu--ko also maintains a positive effect ($+6.32 \to +2.78$). This consistency suggests that these effects are not artifacts of a specific decoding setting, but directionally consistent patterns that recur across certain language combinations (statistical significance and cross-benchmark checks are reported in Appendices~\ref{app:stats} and \ref{app:msvamp}).

\paragraph{Case Study: Maltese.}
Another notable case is Maltese (mt). In Fig.~\ref{fig:heatmap-greedy}, Maltese shows large positive effects with Russian ($+8.62$), Thai ($+6.90$), and Spanish ($+6.61$) skeletons, and all three pairs satisfy nominal significance under McNemar's test (Appendix~\ref{app:stats}). However, these effects are attenuated or even reversed under multi-rollout evaluation (e.g., mt--ru: $-0.23$). Therefore, unlike the directionally consistent cases discussed above, the positive effects observed for Maltese are sensitive to the evaluation protocol and to the benchmark (Appendix~\ref{app:msvamp}), and we do not treat them as evidence of a generalizable positive effect. Since the results in this section alone are insufficient to determine the source of this pattern, we further examine it in Sec.~\ref{sec:analysis}.

\paragraph{Asymmetric Negative Effects}
Surface-level similarities such as shared scripts or geographic proximity do not necessarily lead to positive transfer. Mongolian (mn) and Russian share the Cyrillic script, but differ in word order and morphological structure: Mongolian is an agglutinative language, whereas Russian is fusional. Accordingly, Russian skeletons lead to performance degradation under both greedy decoding ($-5.00$) and multi-rollout evaluation ($-2.65$). Similarly, Uyghur (ug) and Chinese are geographically proximate but structurally different: Uyghur is an SOV agglutinative language, whereas Chinese is an SVO analytic language. Chinese skeletons degrade Uyghur performance ($-6.45$ greedy; $-1.12$ multi-rollout). These cases suggest that surface-level similarity alone is insufficient to predict the direction of skeleton transfer; this negative pattern largely persists on MSVAMP (Appendix~\ref{app:msvamp}).

\paragraph{The Necessity of LASEF.}
Taken together, the effects of skeleton language fall into three patterns: (1) \textit{directionally consistent effects}, where the sign of positive transfer holds under both greedy and multi-rollout evaluation; (2) \textit{evaluation- and benchmark-dependent effects}, where large greedy gains are attenuated under multi-rollout evaluation or on an unseen benchmark; and (3) \textit{asymmetric negative effects}, where negative transfer occurs despite shared surface affinity. These heterogeneous, largely exploratory patterns (Appendices~\ref{app:stats} and \ref{app:msvamp}) show that a single evaluation protocol is insufficient to fully characterize skeleton-language effects, supporting the need for \LASEF as a multi-level empirical analysis framework rather than a rule-based selector.

\section{Analysis}
\label{sec:analysis}

\begin{figure}[!ht]    
    \centering
    \includegraphics[width=\columnwidth]{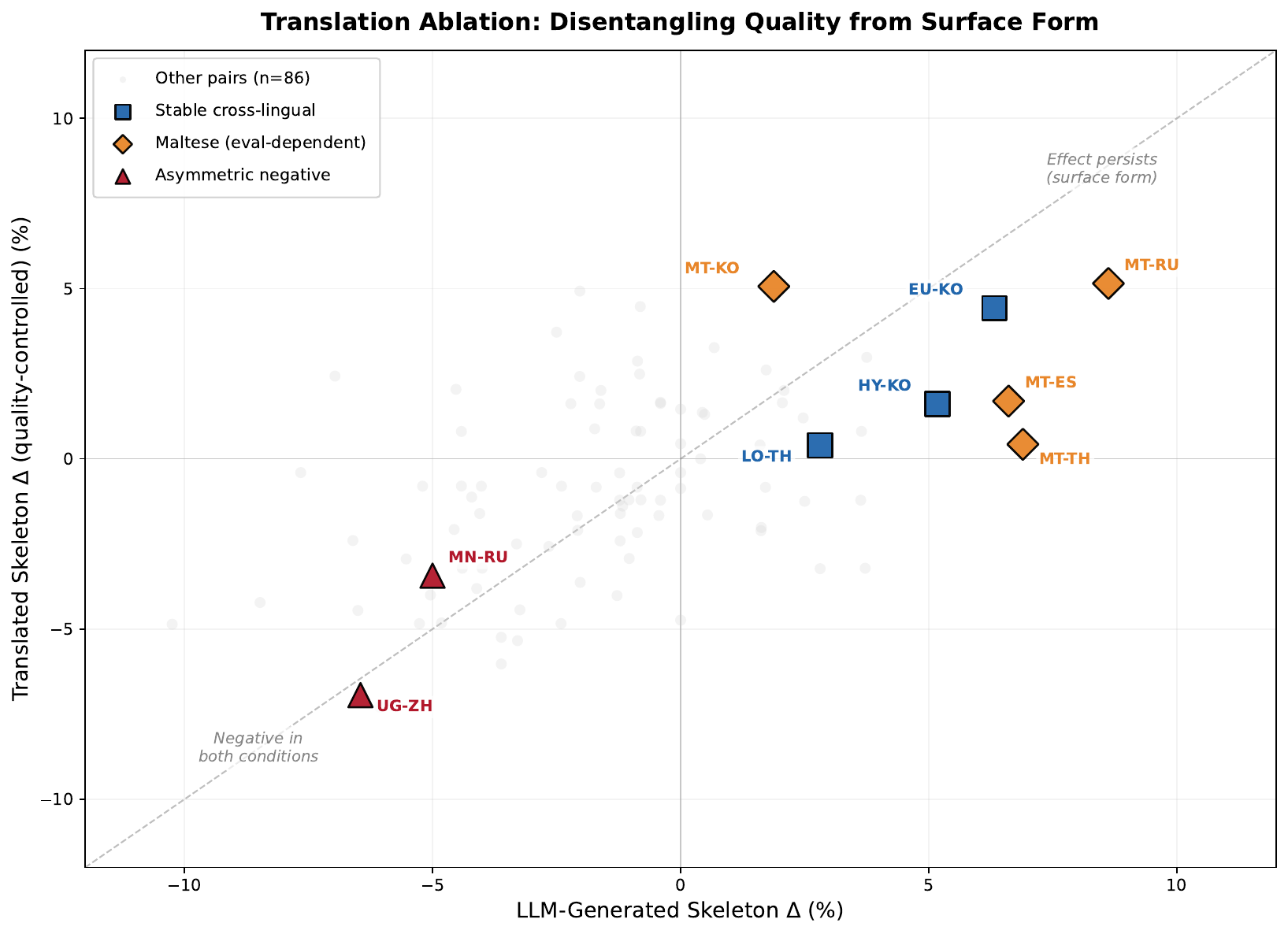}
    \caption{\small Translation ablation results. Each point is a (target, skeleton) pair, with $\Delta$ from LLM-generated skeletons on the x-axis and translated skeletons on the y-axis. Points near the diagonal indicate effects robust to skeleton quality.}
    \label{fig:translation_scatter}
    \vspace{-0.15in}
\end{figure}

\noindent
The effects observed in Sec.~\ref{sec:low-resource-nonen} may conflate two factors: differences in skeleton \textit{generation quality} and \textit{cross-lingual alignment} between skeleton and target language. To disentangle these factors, we conduct a translation ablation. Specifically, we translate the English skeleton into five non-English languages (zh, es, ru, ko, th) using GPT-5-mini, and then run greedy-decoding inference with the translated skeletons, comparing them against the English-skeleton baseline.

Since the semantic content is fixed, any observed difference more directly reflects surface-form variation rather than skeleton generation quality. Fig.~\ref{fig:translation_scatter} visualizes the results: the $x$-axis denotes $\Delta$ with LLM-generated skeletons, the $y$-axis denotes $\Delta$ with translated skeletons, and points near the diagonal ($y{=}x$) indicate effects that are less sensitive to skeleton generation quality. The qualitative LLM-as-a-Judge analysis is provided in Appendix~\ref{Appendix:skeleton_qualitative}.

\paragraph{Directionally Consistent Pairs Persist}
The directionally consistent pairs from Sec.~\ref{sec:low-resource-nonen} fall in the upper-right region of Fig.~\ref{fig:translation_scatter} and retain a positive sign under translation ablation: Basque--Korean ($+6.32 \to +4.42$), Armenian--Korean ($+5.17 \to +1.61$), and Lao--Thai ($+2.81 \to +0.40$). Although the magnitude of $\Delta$ decreases, the positive sign is preserved, suggesting that these effects are not solely attributable to skeleton generation quality.

\paragraph{Maltese: A Fragile Surface Effect}
Maltese is the clearest evaluation-dependent case: large greedy gains with several non-English skeletons are attenuated or reversed under multi-rollout evaluation. The translation ablation shows this is not a generation-quality artifact: with content fixed and only the surface language varied, the gains stay positive (mt--ru: $+5.15$, mt--ko: $+5.06$). It is thus a genuine surface-form effect, but one sensitive to the decoding protocol and to the evaluation benchmark: on MSVAMP, the mt--ru gain reverses (Appendix~\ref{app:msvamp}).

\paragraph{Asymmetric Negatives Persist}
In contrast, the asymmetric negative pairs, Mongolian--Russian ($-3.43$) and Uyghur--Chinese ($-6.94$), lie near the diagonal in the lower-left quadrant of Fig.~\ref{fig:translation_scatter}: both LLM-generated and translated skeletons yield comparable negative transfer, so it is not a quality artifact. These pairs share surface affinity (script or geography) but lack structural alignment (Appendix~\ref{app:typology}), so the effect tracks misalignment, not quality or surface similarity.

Together with the stable and Maltese pairs, this shows that skeleton-language effects are not reducible to generation quality but reflect genuine skeleton--target interactions---effects that cannot be predicted a priori and must be found empirically, the role of \LASEF.

\section{Conclusion}

We analyzed the problem of skeleton language selection in multilingual mathematical reasoning through the Language-Aware Skeleton Exploration Framework (\Lasef), an empirical analysis framework rather than an automatic selector of an optimal skeleton language. Our experiments show that English skeletons are a reasonable default, but their average gains are small and they are not a one-size-fits-all solution. By combining greedy decoding, multi-rollout evaluation, translation ablation, and cross-benchmark validation, we further find that skeleton-language effects can be categorized into three patterns: directionally consistent, evaluation- and benchmark-dependent, and asymmetric negative. These results answer our research question conditionally rather than universally: skeleton language is a context-dependent design variable that cannot be fully explained by generation quality alone, and effective---or risky---language combinations should be identified through multi-level exploration.
\section*{Limitations}
Our study has several limitations. First, we focused exclusively on mathematical reasoning; generalizing findings to domains like commonsense reasoning or code generation requires further validation. Second, our training-free approach ensures broad applicability but precludes potential gains from fine-tuning. Third, we maintained a fixed skeleton structure to isolate language effects, leaving the interplay between skeleton design and language choice unexplored. Finally, while we covered diverse languages, the full \Lasef configuration search remains computationally expensive and is intended as an offline diagnostic rather than a per-deployment step; results from an existing benchmark can help flag risky skeleton candidates but should be revalidated whenever the benchmark, model, or task changes (Appendix~\ref{app:msvamp}). Developing efficient predictive methods for this search remains future work.

\section*{Acknowledgments}
This work was supported by Institute of Information \& Communications Technology Planning \& Evaluation (IITP) grants funded by the Korea government (MSIT) (No. 2022-0-00871, Development of AI Autonomy and Knowledge Enhancement for AI Agent Collaboration; No. RS-2026-25525363, AEGIS: Agentic Experts for Generative-AI Inspection Solution). This work also utilized GPU resources from the ``Advanced GPU Utilization Support Program'' funded by MSIT, Republic of Korea (awarded to KyungTae Lim).
\bibliography{custom}

\clearpage
\appendix
\section{Skeleton}\label{Appendix:about-skeleton}
\subsection{Skeleton Generation Method}
The skeleton generation pipeline proposed in this study is constructed as follows.

\paragraph{Input Configuration}
When generating a skeleton, the model is provided with the problem text together with three few-shot examples. These few-shot examples constrain the output style and abstraction level of the skeleton, ensuring that the skeleton is produced in the specified output language $\ell_s$. Meanwhile, the input question itself is presented in the target language $\ell_t$, and the model generates the skeleton based on this input.

\paragraph{Two Components}
Instead of directly presenting the computational steps or explicit reasoning procedures required to solve the problem, the skeleton is designed to abstract and represent the underlying structural elements and the conceptual problem-solving framework. The generated skeleton is organized into the following two core components:
\begin{enumerate}
    \item \textbf{Problem Structure}: Identifies and defines the \textbf{Objects}, \textbf{Variables}, \textbf{Relationships}, \textbf{Constraints}, and \textbf{Problem Nature} included in the problem.
    \item \textbf{Key Concepts / Tools}: Describes the mathematical/scientific principles and tools essential for solving the problem using standardized academic terminology.
\end{enumerate}

\paragraph{Generation Constraints}
To maintain the level of abstraction and consistency of the skeleton, the generation process is subject to the following constraints. First, the final answer or any specific numerical values are not mentioned explicitly, and the numbers or formulas appearing in the reasoning path are replaced with their conceptual roles (e.g., ``coefficient'', ``boundary value'', ``ratio''). Second, instead of computation-centered verbs such as ``calculate'' or ``compute'' the model is encouraged to use verbs that induce conceptual reasoning, such as ``identify'', ``express'', and ``relate''. Third, while the essential logical structure of the problem-solving process is preserved, peripheral procedural details, redundant descriptions, and errors contained in the reasoning path are excluded. Through these constraints, the skeleton is guided to extract the structural essence of the problem without directly presenting the solution method.

\subsection{System Prompt for Skeleton Generation}
\begin{table*}[p]
\centering
\begin{tcolorbox}[
  width=\textwidth,
  colback=white,
  colframe=black,
  boxrule=0.8pt,
  arc=4pt,
  top=10pt,
  bottom=10pt,
  enhanced,
  title=\bfseries System prompt for skeleton generation,
  colbacktitle=black,
  coltitle=white,
  fonttitle=\normalsize,
  boxed title style={boxrule=0pt},
  titlerule=0pt,
]

\ttfamily\small
Your Role:\\[1em]
You generate minimal, high-level reasoning skeletons for math and science problems.\\

Language Requirement: \\[1em]
Always respond in English, regardless of the language used in the user query. \\[1em]

Goal:\\[1em]
Provide only the *essential structural characteristics* of the problem. You must describe WHAT the problem structure is, never HOW to solve it.\\

Output Format: Provide exactly two sections below.\\

1. **Problem Structure**\\

Identify and characterize:\\[1em]
- \textbf{Objects \& Variables}: Types of entities (quantities, sets, functions, geometric figures), separating constants/parameters from variables. Distinguish knowns vs. unknowns.\\
- \textbf{Relationships}: The nature of connections between entities (e.g., proportional, inverse, functional dependency, spatial). Describe the *type* of relationship, not the derivation.\\
- \textbf{Constraints}: Conditions defining the problem space (equations, inequalities, domain restrictions, boundary conditions, conservation laws).\\
- \textbf{Problem Nature}: (Only if relevant) Briefly note the structural type (e.g., Optimization, Combinatorial, Inverse problem).\\

\textbf{Constraints for this section}:\\[1em]
- Must be problem-specific, not generic.\\
- Must NOT include solution steps, strategies, formulas, or numerical manipulations.\\

2. \textbf{Key Concepts / Tools}\\

List relevant mathematical or scientific principles using standard terminology (e.g., ``Pythagorean Theorem'', ``Conservation of Energy'', ``Bayes' Theorem'').\\

*Constraints for this section:*\\[1em]
- Use established concept names only.\\
- Do NOT explain how to apply them or list formulas.
\end{tcolorbox}
\caption{The prompt instructs the model to produce a concise, high-level outline of the reasoning process, while omitting detailed computations and final answers.}
\label{tab:skeleton_system_prompt}
\end{table*}

The details of the System Prompt applied to the model for skeleton generation are presented in Tab.~\ref{tab:skeleton_system_prompt}. This prompt is designed to ensure that the model does not directly present the final answer or explicit solution procedures (How), but instead abstracts the structural characteristics and core concepts (What) of the problem to construct a conceptual problem-solving plan. The prompt used in the subsequent stage, where the model solves the problem based on the generated skeleton, is presented as follows.

\begin{tcolorbox}[
width=\columnwidth,
  colback=gray!5,
  colframe=black,
  boxrule=0.6pt,
  arc=6pt,
  left=8pt,
  right=8pt,
  top=6pt,
  bottom=6pt,
  title=Solver Prompt for Skeleton
]
\small\raggedright
\textit{Question: [Problem Text]\\
Reasoning Skeleton: [Generated Skeleton]\\[\baselineskip]
Instructions:\\
1. Follow the Reasoning Skeleton above.\\
2. Verify numbers against the Question text.\\
3. Let's think step-by-step. Respond in [Target Language].\\
Final Answer Format: }

\[
\boxed{\text{answer}}
\]

\end{tcolorbox}

\subsection{Few-shot Examples Used for Skeleton Reasoning}
\label{sec:few-shots}

\begin{tcolorbox}[
  colback=blue!3,
  colframe=blue!50!black,
  boxrule=0.5pt,
  arc=4pt,
  left=6pt,
  right=6pt,
  top=4pt,
  bottom=4pt,
  title=Skeleton Few-shot Example 1~(Korean)
]
\small
\textbf{Question:} 삼각형에 길이가 5cm와 12cm인 두 변이 있으며, 그 사이에 직각이 있습니다. 세 번째 변의 길이는 얼마입니까?~
(A triangle has sides of length 5 cm and 12 cm, with a right angle between them. What is the length of the third side?)\\[0.5em]

\textbf{Skeleton:}\\
\textbf{Problem Structure}\\
- Geometric figure: right triangle with three sides\\
- Known quantities: two perpendicular sides (5 cm, 12 cm), right angle constraint\\
- Unknown: third side (hypotenuse) connecting the two legs\\
- Relationship: side lengths satisfy distance relation enforced by right angle property\\[0.3em]
\textbf{Key Concepts / Tools}\\
- Pythagorean theorem\\
- Right triangle geometry
\end{tcolorbox}

\begin{tcolorbox}[
  colback=blue!3,
  colframe=blue!50!black,
  boxrule=0.5pt,
  arc=4pt,
  left=6pt,
  right=6pt,
  top=4pt,
  bottom=4pt,
  title=Skeleton Few-shot Example 2~(Korean)
]
\small
\textbf{Question:} 5명의 후보(Alice, Bob, Charlie, Dave, Eve) 중에서 2명의 위원을 선출해야 합니다. 그러나 Alice와 Bob은 함께 위원이 될 수 없습니다. 위원회를 구성하는 방법은 몇 가지입니까?~(A committee needs to pick 2 members from a group of 5 candidates. However, Alice and Bob cannot be on the committee together. How many ways can the committee be formed?)\\[0.5em]

\textbf{Skeleton:}\\
\textbf{Problem Structure}\\
- Discrete selection problem: choosing k=2 elements from n=5 candidates\\
- Known: total candidates (5), committee size (2), pairwise exclusion constraint\\
- Unknown: number of valid selection configurations\\
- Constraint: pairwise exclusion (Alice and Bob cannot co-occur)\\[0.3em]
\textbf{Key Concepts / Tools}\\
- Combinatorial counting (combinations)\\
- Constraint satisfaction\\
- Complementary counting
\end{tcolorbox}

\begin{tcolorbox}[
  colback=blue!3,
  colframe=blue!50!black,
  boxrule=0.5pt,
  arc=4pt,
  left=6pt,
  right=6pt,
  top=4pt,
  bottom=4pt,
  title=Few-shot Example 3: Rate Problem
]
\small
\textbf{Question:} 기계 A는 8시간 동안 시간당 100개를 생산합니다. 기계 B는 4시간 가동 후 1시간 휴식하고, 그 후 3시간 동안 시간당 120개를 생산합니다. 총 생산량은 얼마입니까?~(Machine A runs for 8 hours producing 100 units/hour. Machine B runs for 4 hours, rests for 1 hour, then runs for 3 hours at 120 units/hour. What is the total production?)\\[0.5em]

\textbf{Skeleton:}\\
\textbf{Problem Structure}\\
- Two independent production processes with different temporal patterns\\
- Variables: production rates, time durations, total output\\
- Machine A: single continuous active interval (8 hours @ 100 units/hour)\\
- Machine B: two disjoint active intervals separated by idle period\\
- Relationship: total output = sum of (rate $\times$ active duration) across all segments\\[0.3em]
\textbf{Key Concepts / Tools}\\
- Rate-time-quantity relationships\\
- Piecewise continuous processes\\
- Linear aggregation
\end{tcolorbox}

For each target language (Korean, Spanish, Chinese, Thai, Swahili), we used few-shot examples where the questions were translated into the respective language, while the skeleton responses were generated in English in all cases.

\subsection{Example of Skeleton Generation}

\begin{tcolorbox}[
  colback=blue!3,
  colframe=gray!60!black,
  boxrule=0.5pt,
  arc=4pt,
  left=6pt,
  right=6pt,
  top=4pt,
  bottom=4pt,
  title=Example~(1): Skeleton Generation
]
\small
\textbf{Question:} 한 자전거가 분당 20피트의 속도로 이동하고 있다. 이 자전거의 속도를 초당 인치로 표현하면 얼마인가?~(A bicycle is traveling at 20 feet per minute. What is the bicycle's speed expressed in inches per second?)\\[0.5em]

\textbf{Skeleton:}\\
\textbf{Problem Structure}\\
- Object: bicycle\\
- Known: speed in feet per minute (20 ft/min)\\
- Unknown: speed in inches per second\\
- Relationships: conversion between units of length (feet to inches) and time (minutes to seconds)\\
- Constraints: 1 foot = 12 inches, 1 minute = 60 seconds\\[0.3em]
\textbf{Key Concepts / Tools}\\
- Unit conversion\\
- Dimensional analysis
\end{tcolorbox}

\begin{tcolorbox}[
  colback=blue!3,
  colframe=gray!60!black,
  boxrule=0.5pt,
  arc=4pt,
  left=6pt,
  right=6pt,
  top=4pt,
  bottom=4pt,
  title=Example~(2): Skeleton Generation
]
\small
\textbf{Question:} 6명의 마녀와 10명의 마법사가 비밀 모임에 참석하였다. 마녀들은 각 마법사와 악수를 하기로 결정했다. 마녀들끼리는 이미 모두 친한 친구이므로 서로 악수를 하지 않고, 마법사들끼리는 서로 싫어해서 악수를 하지 않았다. 이 모임에서 이루어진 총 악수의 수는 몇인가?~(Six witches and ten sorcerers are at an arcane mixer. The witches have decided to shake hands with each of the sorcerers. None of the witches bothered to shake each other's hands, since they are all good friends already, and the sorcerers all sort of hate each other and did not shake hands with other sorcerers. How many handshakes took place at the mixer?)\\[0.5em]

\textbf{Skeleton:}\\
\textbf{Problem Structure}\\
- Discrete interaction problem: handshakes between two distinct groups\\
- Known: number of witches (6), number of wizards (10)\\
- Unknown: total number of handshakes\\
- Relationship: each witch shakes hands with each wizard exactly once\\
- Constraints: no handshakes between witches, no handshakes between wizards\\[0.3em]
\textbf{Key Concepts / Tools}\\
- Combinatorial counting (specifically, Cartesian product of two sets)\\
- Bipartite graph theory (each handshake represents an edge between two disjoint sets)

\end{tcolorbox}

\section{Evaluation Setup}\label{Appendix:Evaluation-setup}

\subsection{Chain-of-Thought Prompt}
We adopt Chain-of-Thought~\citep{cot-neurips2022} as our baseline approach. The prompt is shown below.

\begin{tcolorbox}[
width=\columnwidth,
  colback=gray!5,
  colframe=black,
  boxrule=0.6pt,
  arc=6pt,
  left=8pt,
  right=8pt,
  top=6pt,
  bottom=6pt,
  title=Prompt for CoT
]
\small\raggedright
\textbf{System:}\\
\textit{You are a math problem solver. Solve the given problem.\\[\baselineskip]
At the end, clearly state the final answer in the following format:}
\[
\boxed{\text{final answer}}
\]

\vspace{3pt}
\textbf{User:}\\
\textit{Question: [Problem Text]\\[\baselineskip]
Let's think step by step. Respond in [Target Language].}
\end{tcolorbox}

\subsection{Evaluation Protocol}\label{Appendix:Evaluation-protocol}

\paragraph{Decoding Strategy}
To minimize variability in the reasoning process across all experiments, Greedy Decoding was used. The temperature was set to 0, selecting the token with the highest probability at each generation step.

\paragraph{Exact Match (EM) Evaluation}
Quantitative evaluation was performed using the \texttt{math-verify} library~\citep{mathverify}. The Exact Match (EM) score was calculated based on whether the answer extracted in the `\verb|\boxed{}|' format from the model's final output matched the Ground Truth.

\paragraph{Language Consistency Filtering}
To ensure a fair comparison of reasoning capabilities, we included in the evaluation only those samples whose generated output language matched the target language. In particular, we filtered samples satisfying the condition $\ell_a = \ell_t$, thereby preventing evaluation bias that may arise from code-switching or unintended language mixing. Language identification for the generated responses was performed using the FastText classifier~\citep{fasttext}. Applying this filter to the high-resource setting of Tab.~\ref{tab:combined-main-table} retains 98.8/98.4\% (MGSM), 93.1/93.4\% (MATH-500), and 92.5/92.4\% (PolyMath) of samples for target-language/English-translated queries; retention varies more substantially across low-resource languages, and per-language sample counts together with unfiltered accuracies are reported in Appendix~\ref{app:lowresource-full}.

In addition, to ensure a fair comparison between the baseline and the skeleton-based method (e.g., CoT-$\ell_t$ vs.\ \textsc{+SKELETON}), evaluation was conducted only on samples that satisfied the condition $\ell_q = \ell_t$ in both settings. This filtering ensured that the reasoning performance was compared under identical input–output language conditions.

\paragraph{Intersection-based Pairing.}
For comparisons across skeleton languages in Sec.~\ref{sec:low-resource-nonen}, we adopt intersection-based pairing to ensure fair comparison. Specifically, each (target language, skeleton language) pair is compared against its English-skeleton counterpart only on the subset of items where both settings produce $\ell_a=\ell_t$. To ensure reliable analysis, we further require each target language to retain at least 180 valid samples on average across the five non-English skeleton languages (zh, es, ru, ko, th); target languages with fewer samples are excluded. In total, MGSM-LowResource covers 34 target languages; the 19 that satisfy this criterion form the main analysis set of Sec.~\ref{sec:low-resource-nonen}, and combined with the five non-English skeleton languages they yield the 95 target--skeleton pairs tested in Appendix~\ref{app:stats}.

\paragraph{Language Forcing}
Following the approach proposed by \citet{reason-lang-emnlp2025}, we incorporate the instruction ``Respond in \{$l_t$\}'' along with language-specific trigger phrases. Specifically, we use triggers such as ``좋습니다,'' for Korean, ``Sawa,'' for Swahili, and ``Bien,'' for Spanish to encourage the model to initiate generation in the target language.

\subsection{Mathematical Benchmarks}
In this experiment, to comprehensively evaluate the mathematical reasoning capabilities of multilingual LLMs, we utilized three benchmarks with differing problem difficulties and linguistic characteristics.

\paragraph{MGSM (Multilingual Grade School Math)}
A multilingual benchmark built upon the 250-sample test set of GSM8K~\citep{cobbe2021training}~\citep{shi2022language}, consisting of elementary-level arithmetic problems. It is widely used as a standard metric for measuring basic arithmetic reasoning skills in multilingual environments. In addition to the existing 10 languages (Bengali, Chinese, English, French, German, Japanese, Russian, Spanish, Swahili, Thai), we expanded the evaluation scope to include Korean\footnote{\url{https://huggingface.co/datasets/freewheelin/mgsm_ko}} and various low-resource languages.

\paragraph{MATH-500}
A subset of 500 representative questions selected by topic and difficulty from the high-difficulty mathematics dataset MATH~\citep{hendrycks2measuring}. It covers advanced mathematical fields such as Algebra, Geometry, and Calculus, requiring complex mathematical thinking and problem-solving skills beyond simple calculations. Since MATH is originally an English-only benchmark, we utilized existing high-quality human translations\footnote{\url{https://huggingface.co/datasets/appier-ai-research/Multilingual-MATH-500}, \url{https://huggingface.co/datasets/davidkim205/ko-math-500}, \url{https://huggingface.co/datasets/iapp/math-500-th}} for our multilingual evaluation.

\paragraph{PolyMath}
A recently proposed high-difficulty multilingual math reasoning benchmark~\citep{wang2025polymath}, containing 500 complex problems (125 per difficulty level) across four difficulty levels in 18 languages. Composed of high-quality questions reflecting the cultural and linguistic contexts of each language region rather than simple translations, it is suitable for analyzing the deep multilingual reasoning limitations and performance gaps of LLMs.

\subsection{Models}
We selected \textsc{Qwen2.5} and \textsc{Llama 3.1}, the latest open-source model families with verified multilingual processing capabilities and reasoning performance.

\paragraph{Qwen2.5}~\citep{qwen2025qwen25technicalreport}
Developed by Alibaba Cloud, the \textsc{Qwen2.5} series has demonstrated exceptional performance, particularly in mathematical reasoning and coding. It offers various parameter sizes ranging from 0.5B to 72B and is pre-trained on a vast multilingual corpus, exhibiting strong performance even in non-English languages. In this study, we used the 7B, 14B, and 72B models to analyze the effect of skeletons according to model scale expansion.

\paragraph{Llama 3.1}~\citep{grattafiori2024llama3herdmodels}
As the successor to Meta's \textsc{Llama 3}, \textsc{Llama 3.1} features an extended context window of 128K tokens and enhanced multilingual support. To verify whether the skeleton-based methodology is effective in general-purpose LLMs not specialized for specific domains, we used the 8B and 70B models in our experiments.

\section{Details of MGSM Low-Resource Dataset Construction}
\label{appendix:mgsm-translation}

This section details the construction process and rigorous quality control system of the \textbf{MGSM Low-Resource} benchmark newly established in this study. To expand the existing MGSM dataset to 34 low-resource languages, we introduced a Dual-Model Translation Pipeline applying prompt engineering and a multi-stage verification procedure.

\subsection{Selection Criteria for Low-Resource Languages}
Target languages were selected from ultra-low-resource languages that account for less than $0.05\%$ of the data within the large-scale corpus \texttt{allenai/c4}\footnote{\url{https://huggingface.co/datasets/allenai/c4}}. This was to identify languages that are not sufficiently included in the pre-training data of modern LLMs, leading to underestimated performance or lack of research.

The 34 finally selected languages were chosen considering typological diversity, including various language families, writing systems, and geographical distributions. The detailed list is provided in Tab.~\ref{tab:target-languages}.

\begin{table}[h]
\centering
\small
\resizebox{\columnwidth}{!}{%
\begin{tabular}{p{2.6cm}p{4.6cm}p{2.2cm}}
\toprule
\textbf{Region} & \textbf{Languages} & \textbf{Script} \\
\midrule
Africa & Amharic & Ge'ez \\
\midrule
Europe & Basque, Maltese, Armenian & Latin, Armenian \\
\midrule
Central Asia & Kazakh, Kyrgyz, Uyghur, Pashto & Cyrillic, Arabic \\
\midrule
South Asia & Nepali, Sinhala, Sindhi, Punjabi, Gujarati, Marathi, Odia, Tamil, Malayalam & Indic, Arabic \\
\midrule
SE Asia \& Pacific & Lao, Mongolian & Lao, Cyrillic \\
\bottomrule
\end{tabular}%
}
\caption{List of the 19 low-resource target languages included in the MGSM-LowResource dataset.}
\label{tab:target-languages}
\end{table}

\subsection{Dataset Construction Pipeline}
The dataset construction adopted a dual-model pipeline utilizing Gemini~(\texttt{gemini-3.0-flash})~\citep{geminiteam2025geminifamilyhighlycapable} and GPT~(\texttt{gpt-5-mini})~\citep{openai2024gpt4technicalreport} in parallel to eliminate single-model dependency and maximize translation quality. The entire process consists of three stages: \textbf{Generation-Validation-Selection}.

\subsubsection{Stage 1: Prompt-Based Generation and Numerical Consistency Check}
\label{sec:pre-filtering}

For each problem, two models independently generate translation candidates ($T$). To prevent distortion of mathematical symbols or unnecessary additions and ensure translation accuracy, we designed and applied a system prompt including a professional translator persona and four key constraints (numerical preservation, meaning retention, natural fluency, and output format compliance).

The details of the prompt used are as follows:

\begin{tcolorbox}[ colback=gray!5, colframe=black, boxrule=0.6pt, arc=6pt, left=8pt, right=8pt, top=6pt, bottom=6pt, title=Translation System Prompt ] 
\small
\textit{ "You are a professional translator. Translate the given text to [Target Language]. \\ Rules: \\ 1. Preserve all numbers, mathematical expressions, and proper nouns exactly as they appear \\ 2. Maintain the original meaning and context \\ 3. Use natural, fluent [Target Language] \\ 4. Only output the translation, nothing else" } 
\end{tcolorbox}

Particularly, to guarantee numerical accuracy, which is core information in math word problems, a \textbf{Numerical Preservation} check was performed immediately after generation. We calculated the recall of numerical entities between the source text ($E$) and the translated text ($T$), and if it fell below the threshold ($80\%$), the translation was immediately discarded, and the regeneration process was triggered.

\begin{equation}
    \text{Recall}_{\text{num}}(T, E) < 0.8 \implies \text{Regenerate}
\end{equation}

Through this strict filtering process, numerical errors were preemptively blocked, and only valid candidates adhering to prompt constraints were passed to the next stage.

\subsubsection{Stage 2: Back-Translation Quality Selection (BTQS)}

To select the optimal translation among the candidates ($T_{\text{gemini}}, T_{\text{gpt}}$) that passed the numerical check, we applied the Back-Translation Quality Selection (BTQS) methodology.

\begin{enumerate}
    \item \textbf{Back-Translation}: Each candidate translation is back-translated into English ($E'$) using the model used for its generation.
    \item \textbf{Semantic Similarity Measurement}: To evaluate semantic equivalence between the source ($E$) and back-translated text ($E'$), cosine similarity is calculated using the \texttt{BAAI/bge-m3} encoder.
    \begin{equation}
        s(T) = \cos\big(\phi(E), \phi(E')\big)
    \end{equation}
    \item \textbf{Final Selection}: The translation with the higher similarity score is adopted as the final data.
    \begin{equation}
        T^* = \argmax_{T \in \{T_{\text{gemini}}, T_{\text{gpt}}\}} s(T)
    \end{equation}
\end{enumerate}

\subsection{Quality Control and Verification Results}
\label{subsec:quality_control}

To confirm the reliability of the constructed dataset, we doubly verified linguistic accuracy and logical integrity based on the outputs of the BTQS pipeline.

First, samples with BTQS similarity scores below the threshold ($0.8$) from the previous stage were considered defective and primarily excluded. Instead of simple exclusion, a \textbf{Regeneration} loop was immediately executed for these samples to iteratively correct them until all data met the quality standards.

For the finally selected samples, \textbf{Round-trip Reasoning Validation} was performed to check the preservation of mathematical essence. To maximize efficiency, this was conducted by \textbf{reusing the back-translated text generated during the BTQS evaluation process} without additional translation. That is, English text generated via 34 target languages as pivots was input into the reasoning model (\textsc{Qwen2.5-7B}), and logical consistency was verified by evaluating whether the model correctly derived the answer.

\begin{table}[t]
\centering
\small
\setlength{\tabcolsep}{3.5pt} %
\resizebox{\columnwidth}{!}{%
\begin{tabular}{lrr|lrr}
\toprule
\textbf{Target Lang} & \textbf{Acc (\%)} & \textbf{$\Delta$} & \textbf{Target Lang} & \textbf{Acc (\%)} & \textbf{$\Delta$} \\
\midrule
\textbf{English (Baseline)} & \textbf{90.00} & -- & Quechua (qu) & 89.60  & -0.40\\
Tamil (ta) & 93.60 & +3.60 & Guarani (gn) & 90.00 & 0.00 \\
Gujarati (gu) & 91.60 & +1.60 & Basque (eu) & 90.00 & 0.00 \\
Kannada (kn) & 91.20 & +1.20 & Amharic (am) & 90.00 & 0.00 \\
Khmer (km) & 91.20 & +1.20 & Javanese (jv) & 90.00 & 0.00 \\
Sinhala (si) & 91.20 & +1.20 & Uyghur (ug) & 89.60 & -0.40 \\
Uzbek (uz) & 91.20 & +1.20 & Tajik (tg) & 89.60 & -0.40 \\
Yoruba (yo) & 90.80 & +0.80 & Sundanese (su) & 89.60 & -0.40 \\
Kyrgyz (ky) & 90.80 & +0.80 & Burmese (my) & 89.60 & -0.40 \\
Sindhi (sd) & 90.40 & +0.40 & Georgian (ka) & 89.60 & -0.40 \\
Nepali (ne) & 90.40 & +0.40 & Pashto (ps) & 89.60 & -0.40 \\
Punjabi (pa) & 90.40 & +0.40 & Armenian (hy) & 89.60 & -0.40 \\
Maltese (mt) & 90.40 & +0.40 & Malayalam (ml) & 89.60 & -0.40 \\
Lao (lo) & 90.40 & +0.40 & Malagasy (mg) & 89.60 & -0.40 \\
Somali (so) & 90.00 & 0.00 & Mongolian (mn) & 89.20 & -0.80 \\
Odia (or) & 90.00 & 0.00 & Kazakh (kk) & 89.20 & -0.80 \\
Marathi (mr) & 90.00 & 0.00 & Kurdish (ku) & 89.20 & -0.80 \\
Cebuano (ceb) & 90.00 & 0.00 &  &  &  \\
\bottomrule
\end{tabular}%
}
\caption{Reasoning accuracy of Qwen2.5-7B on MGSM tasks using Round-trip Translation (English $\rightarrow$ Pivot $\rightarrow$ English). We used the target languages as pivots to validate the semantic consistency of the translated dataset. $\Delta$ denotes the performance difference relative to the English baseline.}
\label{tab:round-trip-acc}
\end{table}

The experimental results in Tab.~\ref{tab:round-trip-acc} strongly support the utility of this verification pipeline. Data routed through many pivot languages, including Tamil (93.6\%), recorded reasoning accuracies comparable to or even exceeding the baseline (90.0\%). Even Kazakh, which showed relatively lower performance, maintained a decent accuracy of 89.2\%, suggesting that information and logical structures essential for problem-solving were robustly preserved without loss during the round-trip conversion to the target language.

Of course, this round-trip reasoning accuracy does not perfectly guarantee the grammatical completeness of the target language text itself. The possibility that the high-performance reasoning model self-corrected minor noise in the back-translated text to derive the answer cannot be ruled out. However, in low-resource language environments where securing native evaluators is realistically difficult, this is judged to be the most effective and scalable proxy metric for determining logical information loss.

Detailed statistics of the finally constructed dataset are shown in Tab.~\ref{tab:final-stats}. The dataset consists of a total of 8,500 high-quality samples, with final contributions by model being 62.1\% for Gemini and 37.9\% for GPT.

\begin{table}[h]
\centering
\small
\begin{tabular}{lc}
\toprule
\textbf{Statistic} & \textbf{Value} \\
\midrule
Total Target Languages & 34 \\
Samples per Language & 250 \\
Total Samples & 8,500 \\
Gemini Selection Ratio & 62.1\% \\
GPT Selection Ratio & 37.9\% \\
Avg. BT Similarity Score & 0.96 \\
\bottomrule
\end{tabular}
\caption{Final statistics of the MGSM-LowResource dataset.}
\label{tab:final-stats}
\end{table}

\section{Linguistic Typology and Affinity/Alignment Classification}
\label{app:typology}
\begin{table}[!h]
\centering
\footnotesize
\resizebox{\columnwidth}{!}{%
\begin{tabular}{lllll}
\toprule
Code & Language & Family & Morph. & Order \\
\midrule
\multicolumn{5}{c}{\textit{Target languages}} \\
\midrule
kk & Kazakh    & Turkic            & Agg & SOV \\
ky & Kyrgyz    & Turkic            & Agg & SOV \\
mn & Mongolian & Mongolic          & Agg & SOV \\
ug & Uyghur    & Turkic            & Agg & SOV \\
hy & Armenian  & IE (Armenian)     & Agg$^{\dagger}$ & SOV \\
lo & Lao       & Kra--Dai          & Ana & SVO \\
eu & Basque    & Isolate           & Agg & SOV \\
mt & Maltese   & Afro-Asiatic (Semitic) & Fus & SVO \\
am & Amharic   & Afro-Asiatic (Semitic) & Fus & SOV \\
ne & Nepali    & IE (Indo-Aryan)   & Fus & SOV \\
si & Sinhala   & IE (Indo-Aryan)   & Fus & SOV \\
ps & Pashto    & IE (Iranian)      & Fus & SOV \\
sd & Sindhi    & IE (Indo-Aryan)   & Fus & SOV \\
pa & Punjabi   & IE (Indo-Aryan)   & Fus & SOV \\
gu & Gujarati  & IE (Indo-Aryan)   & Fus & SOV \\
mr & Marathi   & IE (Indo-Aryan)   & Fus & SOV \\
or & Odia      & IE (Indo-Aryan)   & Fus & SOV \\
ta & Tamil     & Dravidian         & Agg & SOV \\
ml & Malayalam & Dravidian         & Agg & SOV \\
\midrule
\multicolumn{5}{c}{\textit{Skeleton languages}} \\
\midrule
en & English   & IE (Germanic)     & Ana/Fus & SVO \\
zh & Chinese   & Sino-Tibetan      & Ana & SVO \\
es & Spanish   & IE (Romance)      & Fus & SVO \\
ru & Russian   & IE (Slavic)       & Fus & SVO \\
ko & Korean    & Koreanic          & Agg & SOV \\
th & Thai      & Kra--Dai          & Ana & SVO \\
\bottomrule
\end{tabular}%
}
\caption{\small Typological features of the 19 target and six skeleton languages. \textbf{Morph.}: dominant morphological type (Ana:
analytic/isolating; Agg: agglutinative; Fus: fusional). \textbf{Order}: basic constituent order. Scripts are listed in
Tab.~\ref{tab:target-languages}. $^{\dagger}$Armenian exhibits agglutinative nominal morphology. These features derive surface affinity and
structural alignment as defined above.}
\label{tab:typology}
\end{table}

To make the notions of \emph{surface affinity} and \emph{structural alignment} used in Sec.~\ref{sec:low-resource-nonen} precise and reproducible, we define them operationally as functions of language-level typological features, listed for all target and skeleton languages in Tab.~\ref{tab:typology}.

\paragraph{Definitions.}
For a target language $t$ and a skeleton language $s$:
\begin{itemize}
\item \textbf{Surface affinity} $\mathrm{Aff}(t,s){=}1$ iff $t$ and $s$ share at least one of: (i) genetic language family, (ii) writing script, or (iii) geographic macro-region; otherwise $0$.
\item \textbf{Structural alignment} $\mathrm{Align}(t,s){=}1$ iff $t$ and $s$ share \emph{both} (i) dominant morphological type (analytic, agglutinative, or fusional) and (ii) basic constituent order (e.g., SOV vs.\ SVO); otherwise $0$.
\end{itemize}
These are \emph{descriptive} categories, not predictors. As shown in Sec.~\ref{sec:low-resource-nonen}, structural alignment does not guarantee positive transfer (e.g., Maltese is alignment-positive yet evaluation-sensitive), and surface affinity does not prevent negative transfer (e.g., ug--zh and mn--ru share affinity but are misaligned and transfer negatively). The classification removes ambiguity from our terminology rather than asserting a causal rule.

\paragraph{Worked examples.}
Under these definitions, lo--th is both affine (Kra--Dai family) and aligned (analytic, SVO); eu--ko is \emph{not} affine but is aligned (agglutinative, SOV); ug--zh and mn--ru are affine (geography or script) but \emph{not} aligned. This matches the patterns reported in Sec.~\ref{sec:low-resource-nonen}.

\section{Computational Cost Analysis}
\begin{figure}[!h]
    \centering
    \includegraphics[width=1\linewidth]{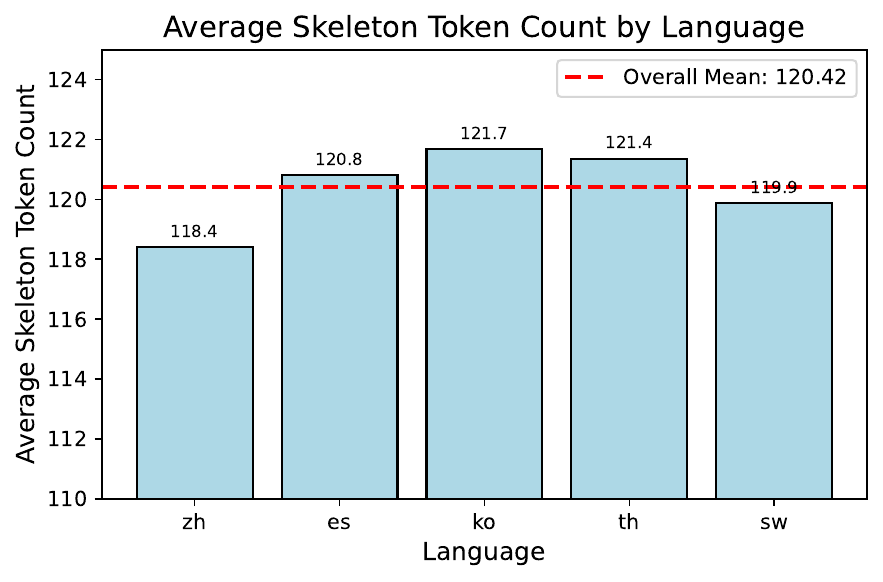}
    \caption{Average token length of skeletons across query languages.}
    \label{fig:skeleton_tokens_vanila_sk}
\end{figure}
\begin{figure}[!h]
    \centering
    \includegraphics[width=1\linewidth]{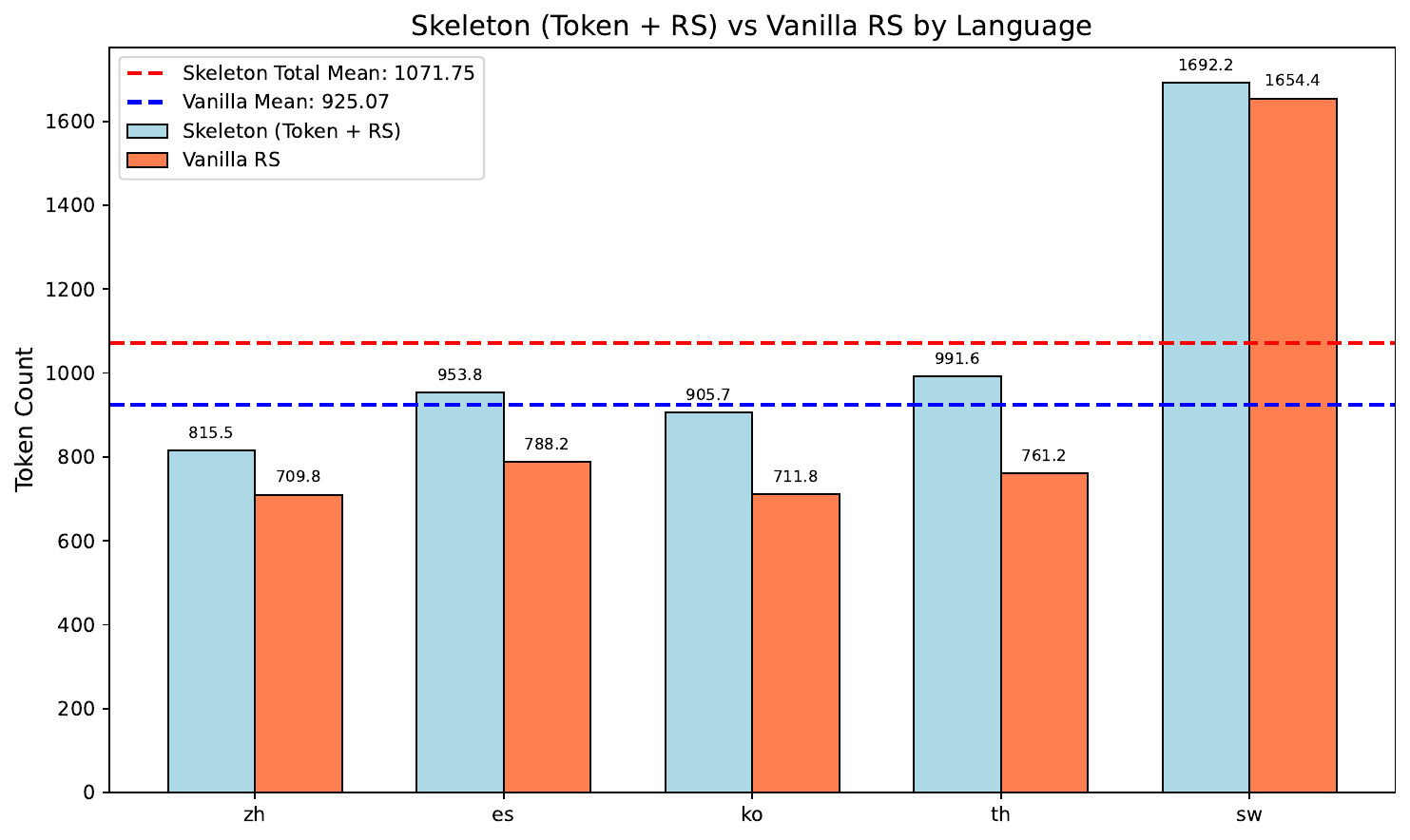}
    \caption{Average generated token counts of the Qwen-series models.}
    \label{fig:skeleton_vs_vanilla}
\end{figure}

Fig.~\ref{fig:skeleton_tokens_vanila_sk}, \ref{fig:skeleton_vs_vanilla} presents the quantitative analysis of the change in computational cost due to the introduction of skeletons. The average length of the skeleton is highly concise, with an overall mean of 120.42 tokens. The average total token count (Skeleton + RS) is 1071.75, representing an increase of approximately 146.68 tokens compared to the vanilla baseline mean of 925.07. Notably, even in languages with high token consumption such as Swahili~(sw), which requires 1692.2 tokens for the full process, the relative overhead remains manageable. These results suggest that the skeleton efficiently guides the model's reasoning process, suppressing unnecessary elaboration or inefficient operations during generation.

\section{Qualitative Analysis}
\label{Appendix:skeleton_qualitative}
\subsection{Quality Assessment of English Skeletons}
\begin{figure}[!ht]    
    \centering
    \includegraphics[width=\columnwidth]{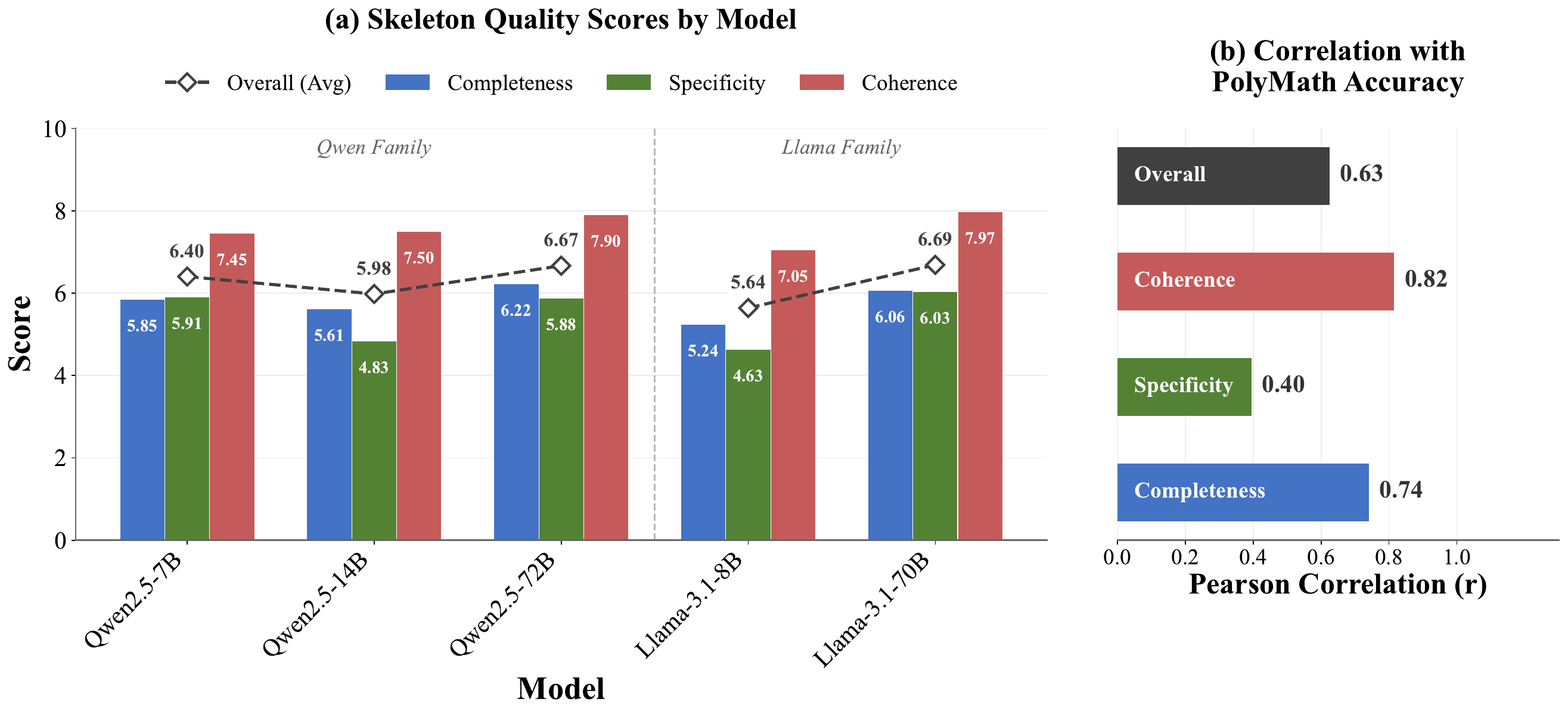}
    \caption{Quality evaluation results for five models in the PolyMath}
\label{fig:llmjudge}
\end{figure}
Fig.~\ref{fig:llmjudge} (Left) shows the change in skeleton generation quality according to model scale. A distinct trend of improved skeleton quality with larger model scale was observed; in particular, large-scale models of 70B or larger recorded a high quality score of 6.7 on average.

Furthermore, the correlation analysis results in Fig.~\ref{fig:llmjudge} (Right) shed light on the relationship between skeleton quality factors and final reasoning performance. `Coherence' ($r=0.82$) and `Completeness' ($r=0.74$) showed strong positive correlations with final performance, whereas `Specificity' ($r=0.40$) showed a relatively lower correlation. This implies that for successful reasoning in the target language, it is more important for the skeleton to include the logical flow of problem-solving and key elements without omission, rather than describing detailed steps specifically.

\subsection{Cross-Lingual Skeleton Quality and the Quality--Performance Gap}
\begin{figure}[!t]    
    \centering
    \includegraphics[width=\columnwidth]{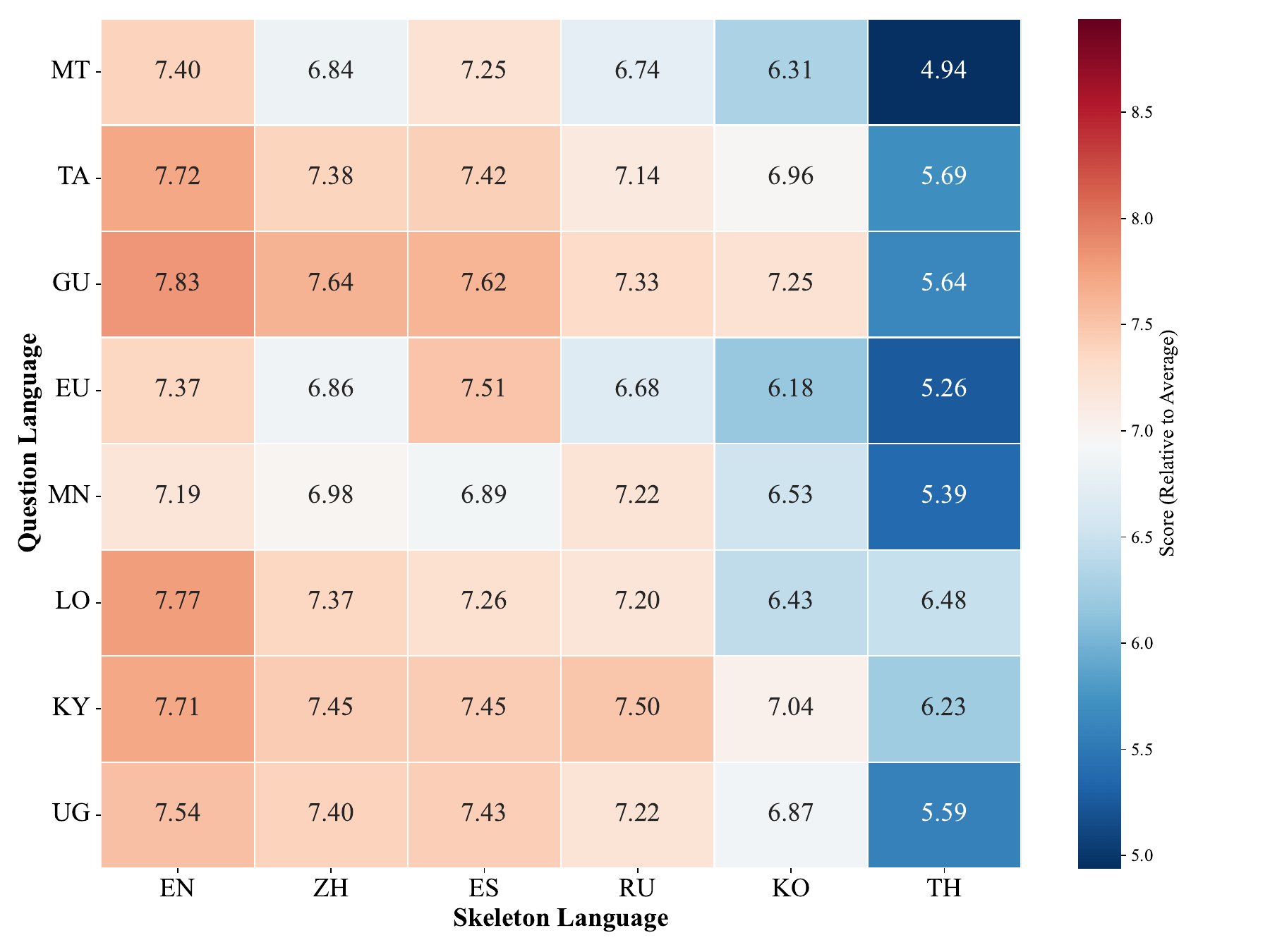}
    \caption{\small Skeleton quality heatmap across language pairs. English skeletons yield highest quality (7.57), while Thai performs worst (5.65).}
    
\label{fig:lang_pair_heatmap}
    \vspace{-0.15in}
\end{figure}
The translation ablation in Sec.~\ref{sec:analysis} controls skeleton \emph{content} to isolate surface form. Here we provide complementary evidence from the opposite direction: we score the \emph{intrinsic generation quality} of skeletons across languages using an LLM-as-a-Judge, and ask whether quality alone can explain the cross-lingual effects in Sec.~\ref{sec:low-resource-nonen}.
\paragraph{English dominates intrinsic quality.}
As shown in Fig.~\ref{fig:lang_pair_heatmap}, English attains the highest average skeleton-quality score ($7.57$), followed by Spanish ($7.35$), Chinese ($7.24$), and Russian ($7.13$), while Korean ($6.69$) and Thai ($5.65$) score lower. This is consistent with English serving as a stable default skeleton language for most targets (Sec.~\ref{sec:low-resource-nonen}).

\paragraph{Quality does not track performance under structural alignment.}
For several structurally aligned pairs, a \emph{lower}-quality skeleton nonetheless yields \emph{higher} performance. For Lao, the Thai skeleton scores $1.29$ points below English in quality ($6.48$ vs.\ $7.77$) yet improves accuracy by $+2.81$; for Basque, the Korean skeleton scores $1.19$ points below English ($6.18$ vs.\ $7.37$) yet improves accuracy by $+6.32$. A quality-only account would incorrectly predict English to be optimal here.

\paragraph{No gap in the absence of structural alignment.}
Conversely, where the target and skeleton share surface affinity but \emph{not} structural alignment, comparable quality coincides with negative transfer. For Uyghur, the Chinese skeleton has quality comparable to English ($7.40$ vs.\ $7.54$) but performance drops by $-6.45$; for Mongolian, the Russian skeleton even slightly exceeds English in quality ($7.22$ vs.\ $7.19$) yet performance drops by $-5.00$. Following the affinity/alignment distinction in Sec.~\ref{sec:low-resource-nonen}, these are precisely the pairs that share surface affinity (script or geography) but differ in morphological type and word order.

\paragraph{Takeaway.}
Together with the translation ablation (Sec.~\ref{sec:analysis}), this indicates that skeleton-language effects are not reducible to generation quality: structural alignment can compensate for lower intrinsic quality, whereas its absence is not offset by high quality. We regard this as supporting evidence for treating skeleton language as a design variable to be identified through exploration, and---consistent with our framing throughout---as an exploratory hypothesis rather than a causal claim.

\subsection{Error Analysis: Impact of Skeleton}
\begin{table}[!ht]
\centering
\resizebox{\columnwidth}{!}{
\begin{tabular}{l l r r r r r}
\toprule
\textbf{Language} & \textbf{Method} 
& \textbf{Problem} 
& \textbf{Conceptual} 
& \textbf{Reasoning} 
& \textbf{Calculation} 
& \textbf{Output Gen.} \\
\midrule
\multirow{2}{*}{kk}
& \textsc{CoT-$\ell_t$}   & 100 & 61 & 129 & 33 & 116 \\
& \textsc{+SKELETON}  &  91 & 54 & 105 & 18 & 135 \\
\midrule
\multirow{2}{*}{ug}
& \textsc{CoT-$\ell_t$}   & 106 & 74 & 156 & 57 &  95 \\
& \textsc{+SKELETON}  & 100 & 60 & 134 & 33 & 135 \\
\midrule
\multirow{2}{*}{mn}
& \textsc{CoT-$\ell_t$}   & 107 & 78 & 157 & 47 &  88 \\
& \textsc{+SKELETON}  & 108 & 62 & 142 & 41 & 115 \\
\midrule
\multirow{2}{*}{hy}
& \textsc{CoT-$\ell_t$}   &  80 & 57 & 108 & 31 & 155 \\
& \textsc{+SKELETON}  &  88 & 50 &  95 & 22 & 154 \\
\midrule
\multirow{2}{*}{lo}
& \textsc{CoT-$\ell_t$}   & 115 & 82 & 159 & 35 & 179 \\
& \textsc{+SKELETON}  & 133 & 75 & 156 & 27 & 123 \\
\midrule
\multirow{2}{*}{eu}
& \textsc{CoT-$\ell_t$}   & 135 & 88 & 160 & 47 &  76 \\
& \textsc{+SKELETON}  & 140 & 85 & 155 & 39 &  97 \\
\midrule
\multirow{2}{*}{mt}
& \textsc{CoT-$\ell_t$}   & 112 & 83 & 154 & 47 &  98 \\
& \textsc{+SKELETON}  & 124 & 66 & 154 & 48 & 119 \\
\midrule
\multirow{2}{*}{ne}
& \textsc{CoT-$\ell_t$}   &  77 & 34 &  93 & 23 & 135 \\
& \textsc{+SKELETON}  &  72 & 44 &  86 & 26 & 139 \\
\bottomrule
\end{tabular}
}
\caption{Comparison of \textsc{CoT-$\ell_t$} and \textsc{+SKELETON} Results Across Different Reasoning Stages}
\label{tab:error_analysis}
\end{table}

To investigate the underlying causes of the performance changes observed in Tab.~\ref{tab:low-resource-mgsm}, we conducted an in-depth error analysis on the entire evaluation dataset using the LLM-as-a-Judge methodology. Tab.~\ref{tab:error_analysis} summarizes the frequency changes of the five key error types before and after applying the skeleton. The detailed evaluation protocol is described in Appendix~\ref{Appendix:Evaluation-protocol}.

To qualitatively analyze the effectiveness of the skeleton, we constructed an automated error classification pipeline using the \texttt{gpt-5-mini} model. For each incorrect sample, the model response was provided as input, and the prompt was designed to identify the most decisive error type among the following five categories:

\begin{enumerate}
    \item \textbf{Problem Comprehension Error}: Misunderstanding the problem's goal, constraints, or key variables.
    \item \textbf{Conceptual Error}: Incorrect application or omission of necessary formulas, theorems, or scientific principles.
    \item \textbf{Reasoning Error}: Errors in logical flow or inclusion of incorrect reasoning steps.
    \item \textbf{Calculation Error}: Mistakes in arithmetic or symbolic operations despite a correct approach.
    \item \textbf{Output Generation Error}: Format errors (missing \verb|\boxed{}|), infinite repetition, language mismatch, etc.
\end{enumerate}

The evaluation model outputs the presence of each error type and a corresponding description in JSON format, which allowed for a quantitative comparative analysis of error patterns by language and condition.

\paragraph{Skeletons as Cognitive Scaffolding}
Analysis of language groups with significant performance improvement (kk, ug, mn, hy) confirmed that the introduction of skeletons significantly suppresses Reasoning Errors and Calculation Errors.
In the case of Kazakh~(kk), which recorded the largest performance gain, Reasoning Errors decreased from 129 to 105 ($-18.6\%$) and Calculation Errors plummeted from 33 to 18 ($-45.5\%$) after applying the skeleton.
This suggests that the English skeleton serves as \textit{cognitive scaffolding} that decomposes complex problems into manageable sub-units, effectively preventing the model from deviating from the logical path or committing simple arithmetic mistakes during the reasoning process.

\begin{table*}[!t]
\centering
\footnotesize
\resizebox{\textwidth}{!}{%
\begin{tabular}{lcccccc>{\columncolor{yellow!15}}c|cccccc>{\columncolor{yellow!15}}c|cccccc>{\columncolor{yellow!15}}c}
\toprule
\multirow{2}{*}{\centering\textbf{Method}} & \multicolumn{7}{c}{\textbf{MGSM}} & \multicolumn{7}{c}{\textbf{MATH-500}} & \multicolumn{7}{c}{\textbf{PolyMath}} \\
\cmidrule(lr){2-8} \cmidrule(lr){9-15} \cmidrule(lr){16-22}
& \textbf{zh} & \textbf{es} & \textbf{ko} & \textbf{th} & \textbf{sw} & \textbf{te} & \textbf{AVG.} & \textbf{zh} & \textbf{es} & \textbf{ko} & \textbf{th} & \textbf{sw} & \textbf{te} & \textbf{AVG.} & \textbf{zh} & \textbf{es} & \textbf{ko} & \textbf{th} & \textbf{sw} & \textbf{te} & \textbf{AVG.} \\

\midrule
\rowcolor{gray!10} \multicolumn{22}{c}{\textbf{\textit{Llama-3.1-8B-Instruct}}} \\
\midrule
\textsc{CoT}$(\ell_q=\text{target})$ &
71.5 & 74.7 & 58.6 & 65.5 & 61.7 & 56.2 & 64.7 &
33.2 & 37.7 & 29.8 & 31.0 & 31.9 & 25.1 & 31.5 &
23.6 & 24.9 & 26.2 & 19.3 & 27.6 & 16.2 & 23.0 \\
+ \textsc{Skeleton}&
68.3 & 75.5 & 61.0 & 65.1 & 63.8 & 57.4 & 65.2 &
34.7 & 35.6 & 34.0 & 32.0 & 34.2 & 25.9 & 32.7 &
24.1 & 26.3 & 24.4 & 21.1 & 29.1 & 16.8 & 23.6 \\
\textsc{CoT}$(\ell_q=\text{EN}^{\dagger})$ &
69.3 & 78.3 & 64.9 & 69.1 & 68.3 & 63.9 & 69.0 &
33.8 & 41.6 & 29.7 & 33.3 & 32.5 & 26.3 & 32.9 &
22.4 & 27.6 & 24.4 & 21.6 & 32.4 & 17.3 & 24.3 \\
+ \textsc{Skeleton}&
69.8 & 73.9 & 63.7 & 70.3 & 66.7 & 62.7 & 67.9 &
37.7 & 37.3 & 32.9 & 31.6 & 38.8 & 29.6 & 34.7 &
26.3 & 28.0 & 27.3 & 22.4 & 33.9 & 17.3 & 25.9 \\
\midrule
\rowcolor{gray!10} \multicolumn{22}{c}{\textbf{\textit{Llama-3.1-70B-Instruct}}} \\
\midrule
\textsc{CoT}$(\ell_q=\text{target})$ &
86.3 & 88.0 & 83.9 & 86.8 & 83.9 & 81.9 & 85.1 &
50.1 & 50.1 & 49.0 & 51.0 & 48.6 & 43.2 & 48.7 &
29.8 & 28.6 & 32.2 & 29.1 & 35.0 & 25.0 & 30.0 \\
+ \textsc{Skeleton}&
88.8 & 81.5 & 82.7 & 87.2 & 78.9 & 80.3 & 83.2 &
50.3 & 45.6 & 50.4 & 52.3 & 49.1 & 45.4 & 48.9 &
30.7 & 29.2 & 31.2 & 30.4 & 33.8 & 26.9 & 30.4 \\
\textsc{CoT}$(\ell_q=\text{EN}^{\dagger})$ & 
85.5 & 88.4 & 84.7 & 85.1 & 83.3 & 83.1 & 85.0 &
49.5 & 50.9 & 51.9 & 47.2 & 51.1 & 46.2 & 49.5 &
29.4 & 32.9 & 32.9 & 28.7 & 36.3 & 26.0 & 31.0 \\
+ \textsc{Skeleton}& 
85.9 & 82.3 & 83.9 & 82.7 & 82.5 & 82.7 & 83.3 &
51.1 & 50.9 & 51.2 & 49.0 & 48.2 & 45.6 & 49.3 &
29.2 & 29.7 & 31.7 & 28.3 & 36.3 & 27.3 & 30.4 \\
\bottomrule
\end{tabular}%
}
\caption{Multilingual mathematical reasoning results for the Llama 3.1 series (8B, 70B).
The experiments use the MGSM, MATH-500, and PolyMath benchmarks. $\ell_q=\text{target}$ indicates that questions are presented in the target language, while $\ell_q=\text{EN}^{\dagger}$ denotes the use of English-translated questions. The skeleton language is fixed to English in all cases ($\ell_s=\text{EN}$). `AVG.' denotes the average over the six languages.}
\label{tab:llama-results}
\end{table*}

\paragraph{Semantic Interference Causing Comprehension Failure}
Conversely, in language groups where performance improvement was marginal or declined (lo, eu, mt, ne), an increase in Problem Comprehension Errors acted as the main failure factor.
For Lao~(lo), while the change in Reasoning Errors was minimal, Problem Comprehension Errors tended to increase from 115 to 133 ($+15.7\%$). This is interpreted as a phenomenon where the English skeleton distorted the core requirements of the original problem due to semantic interference between languages, or the model became overly biased towards the English instructions of the skeleton, missing the context of the original problem.

\section{Llama 3.1 Experimental Results}\label{Appendix:Llama-results}

While the main text presented experimental results centering on the \textsc{Qwen2.5} series, we performed the same experiments on the Meta \textsc{Llama 3.1} series (8B, 70B) to verify the generalizability of the methodology.

Tab.~\ref{tab:llama-results} summarizes the results of the MGSM, MATH-500, and PolyMath benchmarks for the \textsc{Llama 3.1} model. The overall trends appeared similar to the \textsc{Qwen2.5} series:

\begin{itemize}
    \item \textbf{Improvement in Low-Resource Languages}: In the CoT-$en^{\dagger}$ setting of the \textsc{Llama-3.1-8B} model, a performance improvement from 32.5 to 38.8 ($+6.3$) was observed in MATH-500 for Swahili (sw) when skeletons were applied.
    \item \textbf{Diminishing Returns with Scale}: In the \textsc{Llama-3.1-70B} model, the effect of skeletons was minimal, or performance degradation occurred in some languages (es: 88.0 $\rightarrow$ 81.5).
    \item \textbf{Differences by Benchmark Difficulty}: The most consistent improvement effects appeared in MATH-500, while MGSM and PolyMath showed large variations by language.
\end{itemize}

Averaged over the six languages, Llama-3.1-8B shows a small positive skeleton effect, whereas the effect largely disappears or reverses for Llama-3.1-70B (Tab.~\ref{tab:llama-results}): the overall mean skeleton effect is $+0.81$~pp (8B, target-language queries) and $+0.75$~pp (8B, English-translated queries), versus $-0.43$ and $-0.81$~pp for 70B. The skeleton effect therefore varies across model variants and evaluation conditions rather than being uniformly positive; what is shared with Qwen2.5 is the qualitative tendency of the effect to shrink as model size and baseline performance increase (see also Appendix~\ref{app:ministral}). However, since Llama 3.1 has a relatively lower proportion of multilingual data, the baseline performance in low-resource languages was measured lower than that of Qwen2.5. Particularly for Llama-3.1-70B, which already demonstrated high baseline performance in most conditions, the room for additional improvement by skeletons was limited.

\section{Additional Model Family: Ministral}
\begin{table}[!h]
\centering
\small
\resizebox{\columnwidth}{!}{%
\begin{tabular}{lrrr|rr}
\toprule
\textbf{Model} & \textbf{MGSM} & \textbf{MATH-500} & \textbf{PolyMath} & \textbf{AVG.} & \textbf{AVG.$^\ast$} \\
\midrule
Ministral-3B  & $+1.34$ & $+1.79$ & $+0.74$ & $+1.29$ & $+0.38$ \\
Ministral-14B & $-1.10$ & $+0.35$ & $+1.08$ & $+0.11$ & $-0.12$ \\
\bottomrule
\end{tabular}}
\caption{English-skeleton effect $\Delta$ (pp) for the Ministral family under the CoT-$\ell_t$ setting, averaged over the five languages of Tab.~\ref{tab:combined-main-table} (zh, es, ko, th, sw) per benchmark. `AVG.' denotes the average over all five languages; AVG.$^\ast$ excludes Swahili (sw), which retains relatively few valid samples.}
\label{tab:ministral-results}
\end{table}
\label{app:ministral}
To examine whether the scale-dependent pattern is specific to Qwen2.5 and Llama~3.1, we additionally evaluate a third model family, \textsc{Ministral-Instruct} (3B and 14B)~\citep{ministral3}. To keep the comparison consistent, we fix the query condition to the target language (CoT-$\ell_t$) and evaluate both models on the three benchmarks and five languages of Tab.~\ref{tab:combined-main-table}, yielding 30 cells. $\Delta$ denotes the change in accuracy (pp) from adding an English skeleton under the same query condition.

As shown in Tab.~\ref{tab:ministral-results}, Ministral-3B improves by $+1.29$ pp on average, whereas the average effect for Ministral-14B is only $+0.11$ pp; excluding Swahili (sw), which retains relatively few valid samples, the averages are $+0.38$ and $-0.12$ pp, respectively. Across the 30 cells, baseline accuracy is negatively correlated with the skeleton effect (Spearman $\rho=-0.568$, nominal $p=0.0011$). Thus, the English-skeleton effect is not consistently positive in this additional family either, and it varies across model variants and benchmarks. Because we evaluate only two sizes per family, we do not interpret this as a general scaling law; rather, across all three families we observe the same qualitative tendency that the average skeleton effect decreases as model size and baseline performance increase (mean $\Delta$ under CoT-$\ell_t$: Qwen2.5 $+1.73 \to +1.00$ from 7B to 72B; Llama-3.1 $+0.81 \to -0.43$ from 8B to 70B; Ministral $+1.29 \to +0.11$ from 3B to 14B). These results distinguish the \emph{generality of the framework}---the language decomposition $(\ell_q,\ell_s,\ell_a)$ applies to any model---from the \emph{generality of the observed performance effects}, which do not automatically transfer across models.

\section{PolyMath Difficulty-Level Analysis}
\label{app:polymath-difficulty}
\begin{table}[ht]
\centering
\small
\setlength{\tabcolsep}{4pt}
\resizebox{\columnwidth}{!}{%
\begin{tabular}{lcccc}
\toprule
\multirow{2}{*}{\textbf{Model}} & \multicolumn{2}{c}{\textbf{CoT-$\ell_t$}} & \multicolumn{2}{c}{\textbf{CoT-$en^{\dagger}$}} \\
\cmidrule(lr){2-3}\cmidrule(lr){4-5}
 & Base & +\textsc{Skeleton} & Base & +\textsc{Skeleton} \\
\midrule
\textsc{Qwen2.5-7B}  & 4.75 & 4.91~(\textcolor{red}{+0.16}) & 5.12 & 5.82~(\textcolor{red}{+0.70}) \\
\textsc{Qwen2.5-14B} & 6.62 & 8.40~(\textcolor{red}{+1.78}) & 7.13 & 7.56~(\textcolor{red}{+0.43}) \\
\textsc{Qwen2.5-72B} & 9.27 & 9.97~(\textcolor{red}{+0.70}) & 8.92 & 8.96~(\textcolor{red}{+0.04}) \\
\bottomrule
\end{tabular}%
}
\caption{\small Skeleton gains on the two hardest PolyMath subsets (Top + High; 250 of 500 problems), as Exact-Match accuracy (\%) averaged over five languages (zh, es, ko, th, sw). Parentheses show the gain ($\Delta$) from adding the English skeleton. Compared with the full-set gains (Tab.~\ref{tab:combined-main-table}), improvements shrink on the harder subset, consistent with the diminishing effect of skeletons as difficulty increases.}
\label{tab:polymath-difficulty}
\end{table}

PolyMath is organized into four difficulty levels (Low, Medium, High, Top; 125 problems each). To probe the effect of problem difficulty more directly than the full-set average, we restrict evaluation to the two hardest subsets (Top + High; 250 problems) and recompute the English-skeleton gain for the \textsc{Qwen2.5} family (7B, 14B, 72B), averaged over five languages (zh, es, ko, th, sw). Tab.~\ref{tab:polymath-difficulty} reports the results.

Compared with the gains aggregated over all 500 problems, improvements on the harder subset generally shrink: for \textsc{Qwen2.5-7B} the average CoT-$\ell_t$ gain drops from $+1.20$ to $+0.16$, and for \textsc{Qwen2.5-72B} from $+0.90$ to $+0.70$. The mid-sized \textsc{Qwen2.5-14B} is an exception, showing a larger gain ($+1.78$) on the harder subset, suggesting that skeleton guidance can still aid mid-sized models on difficult problems. Overall, the tendency is consistent with our main finding that the benefit of skeletons diminishes as problem difficulty increases.

\section{Statistical Significance Analysis}
\label{app:stats}

\begin{table}[h]
\centering
\small
\begin{tabular}{llrr}
\toprule
Target & Skeleton & $\Delta$ (\%) & $p$ \\
\midrule
\multicolumn{4}{c}{\textit{Positive effects (Non-English skeleton $>$ English skeleton)}} \\
mt & ru & $+8.62$ & 0.001 \\
mt & th & $+6.90$ & 0.008 \\
mt & es & $+6.61$ & 0.041 \\
\midrule
\multicolumn{4}{c}{\textit{Negative effects (English skeleton $>$ Non-English skeleton)}} \\
ug & th & $-10.25$ & 0.002 \\
mn & th & $-8.47$ & 0.016 \\
kk & ko & $-7.66$ & 0.030 \\
\bottomrule
\end{tabular}
\caption{Pairs that reached nominal significance in McNemar's test before correction for multiple comparisons ($p<0.05$). $\Delta$ denotes the accuracy difference between the non-English skeleton and the English skeleton.}
\label{tab:mcnemar}
\end{table}

To assess the statistical reliability of the non-English skeleton effects discussed in Sec.~\ref{sec:low-resource-nonen}, we conducted McNemar's tests. McNemar's test compares paired binary outcomes over the same set of instances, focusing on the transition patterns between correct and incorrect predictions under two conditions. In our analysis, we tested whether item-level correctness differed significantly between the English skeleton and each non-English skeleton. Using the greedy decoding results of \textsc{Qwen2.5-7B}, we examined a total of 95 target--skeleton pairs, corresponding to 19 low-resource target languages and five non-English skeleton languages.

\paragraph{Results.}
Among the 95 pairs, six reached nominal significance before correction for multiple comparisons ($p<0.05$). These include both positive effects, where a non-English skeleton outperformed the English skeleton, and negative effects, where the English skeleton outperformed the non-English skeleton.

The three positive effects were all observed for Maltese (mt), with mt--ru showing the strongest signal ($p=0.001$). In contrast, negative effects were observed for Uyghur--Thai, Mongolian--Thai, and Kazakh--Korean. These results indicate that non-English skeletons can also significantly degrade performance relative to English skeletons in some cases.

The pairs discussed in the main text as directionally consistent cross-lingual effects, namely eu--ko ($p=0.072$), hy--ko ($p=0.361$), and lo--th ($p=0.401$), do not reach nominal significance under McNemar's test. We therefore do not interpret them as effects confirmed by a single statistical test. Instead, we treat them as consistency-based observations, since the direction of the effect remains stable across greedy decoding and multi-rollout evaluation. In other words, McNemar's test examines item-level correctness transitions under a single decoding setting, whereas our robustness analysis evaluates whether the direction of the effect is preserved across evaluation protocols.

For Maltese, the positive effects are nominally significant under greedy decoding, but they are substantially attenuated or even reversed under multi-rollout evaluation. This is consistent with our interpretation in the main text that Maltese represents an evaluation-sensitive case rather than a directionally consistent transfer pattern; the cross-benchmark results in Appendix~\ref{app:msvamp} further show that the Maltese gains are benchmark-dependent.

\paragraph{English Skeleton vs.\ No Skeleton.}
Following the same procedure, we also tested whether English skeletons significantly outperform the no-skeleton CoT baseline on the 19 low-resource MGSM languages (\textsc{Qwen2.5-7B}, CoT-$\ell_t$). The effect was positive for 13 of the 19 languages, with a mean language-level change of $+2.38$ pp. After FDR correction, however, significant improvements remained for only two languages: Kazakh (kk; $40.08 \to 51.42$, $\Delta=+11.34$, $p_{\mathrm{FDR}}=0.014$) and Uyghur (ug; $27.13 \to 37.65$, $\Delta=+10.52$, $p_{\mathrm{FDR}}=0.014$). The $-5.65$ pp decrease for Maltese was nominally significant before correction ($p=0.049$) but not after FDR correction ($p_{\mathrm{FDR}}=0.307$), and no other language showed a significant difference after correction. When the 19 languages are pooled at the item level ($n=4{,}628$), the overall effect is significant ($\Delta=+2.27$ pp, $p=0.0014$). In other words, the benefit of English skeletons is real but too small to detect at the individual-language level, and it is driven by a few languages such as kk and ug.

\section{Cross-Benchmark Generalization on MSVAMP}
\label{app:msvamp}
\begin{table}[!ht]
\centering
\small\resizebox{\columnwidth}{!}{
\begin{tabular}{llrrl}
\toprule
\textbf{Target} & \textbf{Skel.} & \textbf{MGSM $\Delta$} & \textbf{MSVAMP $\Delta$} & \textbf{Outcome} \\
\midrule
hy & ko & $+5.17$  & $+2.75$ & Retained \\
lo & zh & $+3.63$  & $+2.63$ & Retained \\
lo & th & $+2.81$  & $+2.42$ & Retained \\
eu & ko & $+6.32$  & $0.00$  & Disappeared \\
mt & es & $+6.61$  & $+2.00$ & Attenuated \\
mt & th & $+6.90$  & $+0.43$ & Attenuated \\
mt & ru & $+8.62$  & $-1.27$ & Reversed \\
ug & th & $-10.25$ & $-2.14$ & Retained (neg.) \\
mn & th & $-8.47$  & $-2.53$ & Retained (neg.) \\
kk & ko & $-7.66$  & $-4.63$ & Retained (neg.) \\
\bottomrule
\end{tabular}
}
\caption{Cross-benchmark transfer of skeleton-language effects from MGSM to MSVAMP (\textsc{Qwen2.5-7B}, CoT-$\ell_t$). $\Delta$: accuracy difference (pp) between the non-English and the English skeleton. Pairs were fixed using only the MGSM results before inspecting MSVAMP. Across all 35 target--skeleton cells, the sign is preserved in 23/35 (66\%).}
\label{tab:msvamp-transfer}
\end{table}

To test whether skeleton-language effects observed on MGSM generalize to an unseen test set, we evaluate on MSVAMP~\citep{chen2023breaking}, a separate benchmark whose problems do not overlap with MGSM (up to 1{,}000 mathematical reasoning problems per language). Since MSVAMP natively supports only ten languages, we obtained additional low-resource target languages using the same translation and verification pipeline described in Appendix~\ref{appendix:mgsm-translation}, extending it to eight languages (ko, eu, hy, lo, mn, ug, mt, and kk). We use the seven target languages for which MGSM-LowResource results are available (eu, hy, lo, mn, ug, mt, and kk).

\paragraph{Protocol.}
To rule out post-hoc selection, the evaluation targets, selection rule, and skeleton candidates were fixed using only the MGSM results before inspecting MSVAMP. We selected eu, hy, and lo (directionally consistent pairs in Sec.~\ref{sec:low-resource-nonen}), mn and ug (asymmetric-negative cases), and mt and kk (nominally significant results in Appendix~\ref{app:stats}). For each target language, we applied the five non-English skeleton languages (zh, es, ru, ko, and th) under the target-language query condition and computed $\Delta$, the accuracy difference relative to the English skeleton, on both benchmarks, yielding $7 \times 5 = 35$ target--skeleton cells.

\paragraph{Results.}
Across the 35 cells, effect sizes on MGSM and MSVAMP show a moderate positive rank correlation (Spearman $\rho=0.383$, nominal $p=0.023$), and the sign of the effect is preserved in 23 of the 35 cells (66\%; binomial $p=0.090$). Tab.~\ref{tab:msvamp-transfer} summarizes the key pairs. Two observations stand out. First, the asymmetric-negative pattern transfers relatively well: among the 15 cells for mn, ug, and kk, 14 remain negative on MSVAMP, with only ug--ru reversing ($-4.11 \to +1.79$). Second, the positive Maltese effects are not robust: mt--es remains positive ($+6.61 \to +2.00$), but mt--th largely diminishes ($+6.90 \to +0.43$) and mt--ru reverses ($+8.62 \to -1.27$). The transfer of positive selections is mixed: hy--ko, lo--zh, and lo--th retain their positive direction, whereas eu--ko vanishes to $0.00$.

\paragraph{Implications.}
We therefore do not claim that the optimal skeleton language can be specified in advance for an arbitrary target language. Within the evaluated subset, results from an existing benchmark are more useful for identifying potentially risky skeleton candidates and prioritizing subsequent evaluations than for directly determining an optimal skeleton language. This asymmetry is an exploratory observation, and separate validation remains necessary for a new benchmark, model, or task. As MGSM and MSVAMP focus on elementary arithmetic, testing transfer on harder benchmarks is left for future work.

\section{Model Performance Across Other Low-Resource Languages}\label{app:lowresource-full}
\subsection{Results on Telugu}
\begin{table}[!t]
\centering
\small
\resizebox{\columnwidth}{!}{
\begin{tabular}{lccc}
\toprule
\textbf{Method} & \textbf{MATH-500 (te)} & \textbf{PolyMath (te)} & \textbf{MGSM (te)} \\
\midrule
\rowcolor{gray!10} \multicolumn{4}{c}{\textbf{Qwen2.5-7B-Instruct}} \\
\midrule
CoT-$\ell_t$ & 29.8 & 12.2 & 30.40 \\
\textsc{+SKELETON} & 27.8 & 12.4 & 33.60 \\
CoT-$en^{\dagger}$ & 35.7 & 17.3 & 60.40 \\
\textsc{+SKELETON} & 33.9 & 17.7 & 54.00 \\
\midrule
\rowcolor{gray!10} \multicolumn{4}{c}{\textbf{Qwen2.5-14B-Instruct}} \\
\midrule
CoT-$\ell_t$ & 43.4 & 18.6 & 54.80 \\
\textsc{+SKELETON} & 45.9 & 19.8 & 55.20 \\
CoT-$en^{\dagger}$ & 45.1 & 24.2 & 71.89 \\
\textsc{+SKELETON} & 43.0 & 22.2 & 73.90 \\
\midrule
\rowcolor{gray!10} \multicolumn{4}{c}{\textbf{Qwen2.5-72B-Instruct}} \\
\midrule
CoT-$\ell_t$ & 53.2 & 23.7 & 65.20 \\
\textsc{+SKELETON} & 55.4 & 27.4 & 68.00 \\
CoT-$en^{\dagger}$ & 52.4 & 28.2 & 78.40 \\
\textsc{+SKELETON} & 55.2 & 28.8 & 82.80 \\
\midrule
\rowcolor{gray!10} \multicolumn{4}{c}{\textbf{Llama-3.1-8B-Instruct}} \\
\midrule
CoT-$\ell_t$ & 25.8 & 16.2 & 56.40 \\
\textsc{+SKELETON} & 25.0 & 16.8 & 57.60 \\
CoT-$en^{\dagger}$ & 26.7 & 17.3 & 63.60 \\
\textsc{+SKELETON} & 30.1 & 17.3 & 62.80 \\
\midrule
\rowcolor{gray!10} \multicolumn{4}{c}{\textbf{Llama-3.1-70B-Instruct}} \\
\midrule
CoT-$\ell_t$ & 43.6 & 25.0 & 82.00 \\
\textsc{+SKELETON} & 45.9 & 26.9 & 80.40 \\
CoT-$en^{\dagger}$ & 46.7 & 26.0 & 83.20 \\
\textsc{+SKELETON} & 46.1 & 27.3 & 82.80 \\
\bottomrule
\end{tabular}
}
\caption{Telugu (te) Accuracy (\%) on MATH-500, PolyMath, and MGSM}
\label{tab:te-results}
\end{table}

Tab.~\ref{tab:te-results} presents the performance changes when applying the skeleton to Telugu.

\begin{figure*}
    \centering
    \includegraphics[width=0.75\linewidth]{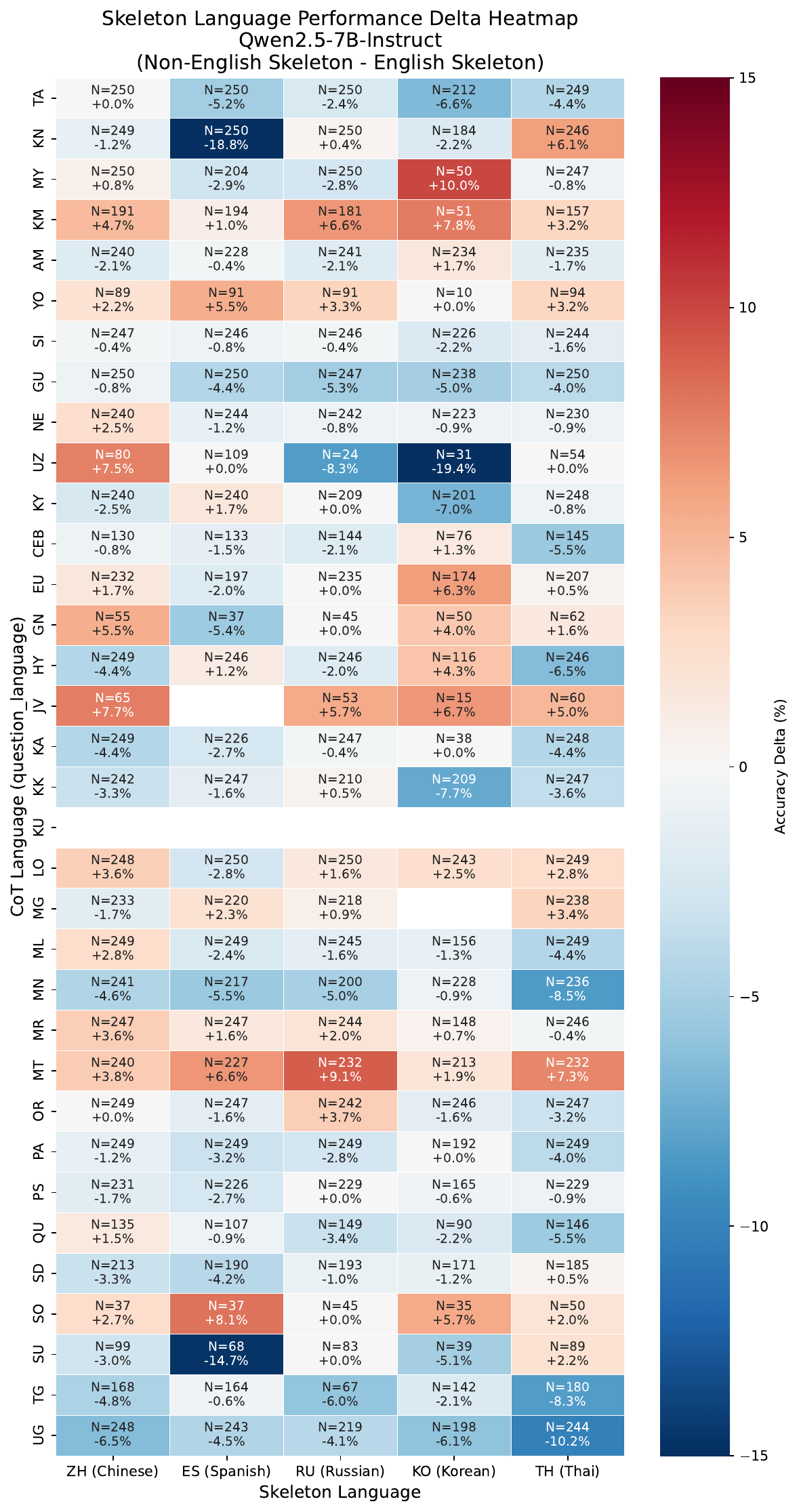}
    \caption{\small Performance differences (Non-English − English) under the CoT-$\ell_t$ setting using \textsc{Qwen2.5-7B} across all 34 target languages. While English generally serves as a stable default skeleton language, several target languages exhibit gains with specific Non-English skeletons, indicating that skeleton effectiveness is language-dependent.
}
\label{fig:all-languages-heatmap}
\end{figure*}
\subsection{Results on All Low-Resource Languages}
Although our experiments cover 34 languages in total, target-language CoT (\textsc{CoT-$\ell_t$}) fails to operate reliably for a subset of languages, making fair comparison difficult. We therefore focus our main analysis on 19 languages where target-language reasoning is stable. For completeness and transparency, results for all 34 languages are reported in Fig.~\ref{fig:all-languages-heatmap}.

\end{document}